\pdfoutput=1

\documentclass[11pt,table]{article}

\usepackage[preprint]{acl}

\usepackage{times}
\usepackage{latexsym}

\usepackage[T1]{fontenc}

\usepackage[utf8]{inputenc}

\usepackage{microtype}

\usepackage{inconsolata}

\usepackage{graphicx}

\usepackage{colortbl}
\usepackage{amsmath, amssymb, bm}
\usepackage[capitalise]{cleveref}
\usepackage{booktabs}
\usepackage{multirow}
\usepackage{amsmath}

\usepackage{paralist}
\usepackage{soul}
\newcommand{\red}[1]{\textcolor{red}{#1}}

\definecolor{dullgreen}{RGB}{102, 153, 102}

\newcommand{\dg}[1]{\textcolor{dullgreen}{#1}}
\definecolor{Highlight}{rgb}{0.89,0.89,0.94}
\definecolor{icecream}{HTML}{F5E7CF}
\newcommand{\ice}{\rowcolor{icecream}}
\definecolor{change}{HTML}{E8C0BB}
\definecolor{ocean}{HTML}{ADD4DD}
\usepackage{array}

\usepackage{algorithm}
\usepackage{algorithmic}
\usepackage{subcaption}

\usepackage{amsmath}
\usepackage{booktabs}
\usepackage{array}
\usepackage{graphicx}

\usepackage{booktabs}
\usepackage{multirow}
\usepackage{amsmath} 
\usepackage{adjustbox}

\usepackage{mdframed}

\usepackage{amsmath}
\usepackage{booktabs}
\usepackage{array}
\usepackage{float}
\usepackage{hyperref}
\usepackage{adjustbox}
\usepackage{enumitem} 

\definecolor{llmevalcolor}{gray}{0.95}

\newcolumntype{C}[1]{>{\centering\arraybackslash}m{#1}}

\newcommand{\q}[1]{q(#1)}
\newcommand{\qcond}[2]{\q{#1 | #2}}
\newcommand{\p}[1]{p_\theta(#1)}
\newcommand{\pcond}[2]{\p{#1 | #2}}

\newcommand{\z}[1]{\boldsymbol{z}_{#1}}
\newcommand{\ttau}[1]{\tau_{#1}}
\newcommand{\s}[1]{s_{#1}}

\newcommand{\dt}{\Delta t}

\newcommand{\zhat}{\hat{\boldsymbol{z}}_0}

\newcommand{\loss}{\mathcal{L}}
\newcommand{\lossrm}[1]{\loss_\mathrm{#1}}
\newcommand{\lvlb}{\lossrm{VLB}}
\newcommand{\lz}{\loss_z}
\newcommand{\ltau}{\loss_\tau}
\newcommand{\lours}{\loss}
\newcommand{\kl}[2]{\mathbb{KL} (#1 \Vert #2)}

\newcommand{\smax}{s_\mathrm{max}}

\newcommand\dif{\mathop{}\!\mathrm{d}}
\newcommand{\bracket}[1]{\left( #1 \right)}
\newcommand{\func}[2]{#1 \bracket{#2}}
\newcommand{\funcup}[2]{\mathrm{#1} \bracket{#2}}

\newcommand{\normal}[1]{\func{\mathcal{N}}{#1}}
\newcommand{\I}{\mathbf{I}}

\newcommand{\R}{\mathbb{R}}

\newcommand{\expect}[1]{\mathbb{E} \left[ #1 \right]}
\newcommand{\var}[1]{\mathbb{V} \left[ #1 \right]}

\newcommand{\norm}[1]{\left\lVert #1 \right\rVert^2}

\newcommand{\lossup}[1]{\loss_{\mathrm{#1}}}

\newcommand{\lanchor}{\lossup{anchor}}

\newcommand{\zvec}{\bm{z}}

\newcommand{\dgs}{\mathrm{DGS}}

\newcommand{\dgsmax}{\dgs_{\mathrm{MAX}}}

\newcommand{\zfpred}{\zvec_{\theta}}

\newcommand{\taupred}{\tau_{\theta}}

\newcommand{\stpt}{\func{s'}{t}}

\newcommand{\lambdat}{\func{\lambda}{t}}
\newcommand{\sigmat}{\func{\sigma}{t}}

\title{Unifying Continuous and Discrete Text Diffusion \\with Non-simultaneous Diffusion
Processes}

\author{
  Bocheng Li\footnotemark[1]$^{1,2}$, 
  Zhujin Gao\footnotemark[1]$^{1,2}$,
  Linli Xu\footnotemark[2]$^{1,2}$      
  \\ $^{1}$School of Computer Science and Technology, University of Science and Technology of China
  \\ $^{2}$State Key Laboratory of Cognitive Intelligence
  \\ \texttt{\{bcli,gaozhujin\}@mail.ustc.edu.cn}, \texttt{linlixu@ustc.edu.cn}
}

\begin{document}
\maketitle
\begingroup 
\renewcommand\thefootnote{\fnsymbol{footnote}} 
\footnotetext[1]{Equal contribution.}          
\footnotetext[2]{Corresponding author.}        
\endgroup
\begin{abstract}
Diffusion models have emerged as a promising approach for text generation, with recent works falling into two main categories: discrete and continuous diffusion models. Discrete diffusion models apply token corruption independently using categorical distributions, allowing for different diffusion progress across tokens but lacking fine-grained control. Continuous diffusion models map tokens to continuous spaces and apply fine-grained noise, but the diffusion progress is uniform across tokens, limiting their ability to capture semantic nuances. To address these limitations, we propose \textbf{\underline{N}}on-simultan\textbf{\underline{e}}ous C\textbf{\underline{o}}ntinuous \textbf{\underline{Diff}}usion Models (NeoDiff), a novel diffusion model that integrates the strengths of both discrete and continuous approaches. NeoDiff introduces a Poisson diffusion process for the forward process, enabling a flexible and fine-grained noising paradigm, and employs a time predictor for the reverse process to adaptively modulate the denoising progress based on token semantics. Furthermore, NeoDiff utilizes an optimized schedule for inference to ensure more precise noise control and improved performance. Our approach unifies the theories of discrete and continuous diffusion models, offering a more principled and effective framework for text generation. Experimental results on several text generation tasks demonstrate NeoDiff's superior performance compared to baselines of non-autoregressive continuous and discrete diffusion models, iterative-based methods and autoregressive diffusion-based methods. These results highlight NeoDiff's potential as a powerful tool for generating high-quality text and advancing the field of diffusion-based text generation.
\end{abstract}

\section{Introduction}

Diffusion models have demonstrated remarkable success in generating high-quality samples in various domains, including vision~\citep{dhariwal2021diffusion, nichol2021improved, ho2021classifier, rombach2022high} and audio ~\citep{chen2020wavegrad, kong2020diffwave}.
Inspired by their achievements, there has been a growing interest in applying diffusion models to text generation tasks~\citep{li2022diffusion, gong2022diffuseq, gao-etal-2024-empowering, zheng2023reparameterized}. 

The core idea behind diffusion models is to corrupt the data through a forward process and then learn to reverse this process to generate new samples. In text generation, existing diffusion models can be broadly categorized into two classes: discrete and continuous diffusion models. Discrete diffusion models treat tokens as discrete random variables and perform state transitions independently for each token using a categorical distribution. While straightforward, this approach fails to capture the continuous and fine-grained nature of language, limiting the potential benefits of multi-step generation. Continuous diffusion models, on the other hand, operate in a continuous space by mapping tokens to continuous representations, enabling more fine-grained perturbations. However, these models typically apply diffusion at the sentence level, resulting in uniform noise levels across all tokens within a sentence, restricting the model's ability to leverage contextual information and recover tokens with varying noise levels based on the surrounding context~\citep{chen-etal-2023-cheaper,wu2024ar}.

\begin{figure*}[t]
    \centering
    \includegraphics[width=1\textwidth]{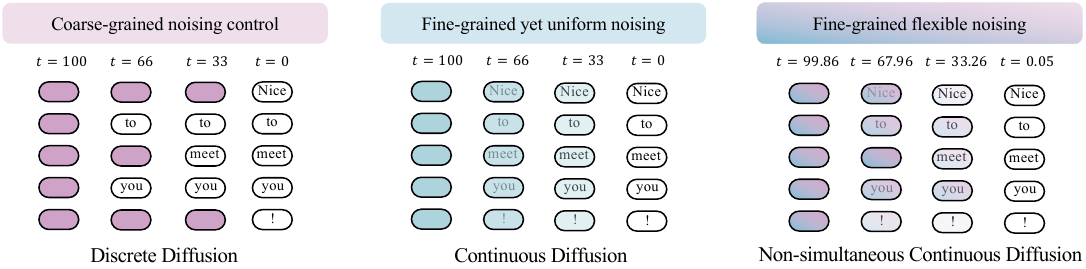}
    \caption{Comparison of the noising paradigms employed by Non-simultaneous Continuous Diffusion and two other diffusion models. The color intensity on the text tokens represents the token-level noising progress (intrinsic time $\tau$). Discrete diffusion applies an independent but coarse-grained noising paradigm to each token within a sentence. In contrast, continuous diffusion utilizes a fine-grained noising schedule but applies it uniformly across all tokens. NeoDiff distinguishes itself by assigning an independent, fine-grained intrinsic time $\tau$ to each token, with finer noising schedule in extrinsic time $t$.}
    \label{fig:fig1}
\end{figure*}

To address these limitations, we propose integrating the complementary strengths of discrete and continuous diffusion approaches, enabling fine-grained noise control at the token level.  This unified approach aims to provide precise token-level control while maintaining continuous-valued noise distributions, which is absent in existing frameworks. While recent text diffusion models~\citep{han2023ssd,gong2023diffuseq, wu2024ar} have made advances, they do not fully address this requirement, necessitating a unified theoretical framework that bridges discrete and continuous diffusion paradigms through a carefully designed forward process. 

Furthermore, we observe that 
existing approaches
primarily focus on enhancing the forward process, overlooking the inherent varying difficulties in denoising different tokens and the impact of generation context. 
In analyzing the reverse process, we recognize that tokens with lower noise levels can guide the recovery of more heavily corrupted tokens, thereby enhancing the overall text generation quality.

In response to these challenges, we present \textbf{\underline{N}}on-simultan\textbf{\underline{e}}ous C\textbf{\underline{o}}ntinuous \textbf{\underline{Diff}}usion Models (NeoDiff), which unifies discrete and continuous diffusion models through a bi-temporal framework. The key insight is to generalize the time variable in previous diffusion models into extrinsic time $t$, representing the diffusion progress of the entire sentence, and intrinsic time $\tau$, tracking the diffusion progress of each individual token. This generalization enables us to introduce a novel Poisson process as the forward process, seamlessly integrating the flexibility of discrete noise with the fine granularity of continuous noise. An overview of this noising paradigm is illustrated in Figure~\ref{fig:fig1}.

To optimize the reverse process, we develop a context-aware time predictor that estimates the intrinsic time $\tau$ using an adaptive modulation function to guide the denoising process. The extrinsic time schedule is further calibrated through Bayesian optimization, providing precise control over the noise distribution.

NeoDiff achieves a fine-grained, improved diffusion process in both forward and reverse directions, naturally overcoming the constraints of previous discrete and continuous diffusion models, and exhibiting superior generation quality. We evaluate NeoDiff on a diverse set of NLP tasks, including machine translation, paraphrasing, text simplification, and question generation. NeoDiff consistently outperforms previous non-autoregressive diffusion-based and iteration-based methods, as well as autoregressive diffusion baselines. Specifically, our contributions can be summarized as follows:

\begin{itemize}
\item We introduce NeoDiff, a unified theoretical framework that combines the advantages of discrete and continuous noise, generalizing and unifying existing text diffusion models.
\item We propose the Poisson diffusion process as the forward process, enabling fine-grained corruption of text data, a context-aware time predictor that adaptively modulates the reverse process based on semantic context, and an optimized extrinsic time schedule for precise noising control.
\item  We conduct extensive experiments to evaluate the effectiveness of NeoDiff and compare it to existing text diffusion models. Our results highlight the advantages of our unified framework and suggest its potential to advance diffusion-based text generation.
\end{itemize}

\section{Background}\label{sec:background}
\subsection{Diffusion Models}

Diffusion models assume a gradual noise injection process over time for data samples $\boldsymbol{z}_{0} \in \R^{N \times d}$. The forward diffusion process forms a series of latent variables $\boldsymbol{z}_{1}, \boldsymbol{z}_{2}, \cdots, \boldsymbol{z}_{T}$ satisfying the Markov property, and finally become pure Gaussian noise $\boldsymbol{z}_T \sim \normal{\mathbf{0}, \I}$: 
\begin{align}
    \qcond{\boldsymbol{z}_t}{\boldsymbol{z}_{t - 1}} &= \normal{
        \boldsymbol{z}_t;
        \sqrt{\alpha_t} \boldsymbol{z}_{t - 1},
        \beta_t \I
    },
\end{align}
where $\alpha_t + \beta_t = 1$, determining the degree of noising at time $t$ and consituting the noise schedule. The reverse process is parameterized as 
\begin{align}
    \pcond{\boldsymbol{z}_{t - 1}}{\boldsymbol{z}_t} &= \normal{
        \boldsymbol{z}_{t - 1};
        \func{\mathbf{\mu}_{\theta}}{\boldsymbol{z}_t, t},
        \func{\mathbf{\Sigma}_{\theta}}{\boldsymbol{z}_t, t}
    },
    \label{equ:p}
\end{align}
Here, $\func{\mathbf{\mu}_{\theta}}{\cdot}$ and $\func{\mathbf{\Sigma}_{\theta}}{\cdot}$ are the model's estimates of the distribution mean and covariance matrix, respectively. The training objective is derived from the variational lower bound of the negative log-likelihood loss, and can be then simplified as an MSE loss~\citep{ho2020denoising,li2022diffusion,gao-etal-2024-empowering}:
\begin{equation*}
    \lvlb = \expect{
        \norm{\func{\z{\theta}}{\z{t}, t} - \z{0}}
        -\log p(\mathbf{z}_0\mid\mathbf{z}_1)
    }
\end{equation*}

\subsection{Discrete Diffusion Models}

Discrete diffusion models directly model the noise on categorical distributions, discarding the assumption that the noise in latent variables follows a normal distribution in continuous space. These models typically represent data as sequences of one-hot vectors and employ a transition matrix to add noise to the data. Among them, \citet{hoogeboom2021argmax} proposed a multinomial diffusion model that employs a uniform noising method. \citet{austin2021structured} introduced D3PM, which employs a noising method with an absorbing state. Specifically, they added an absorbing state [MASK] to the vocabulary, which can only be entered but not exited. The remaining states, at each diffusion step, either stay in the current state or enter the absorbing state with a certain probability. Recently, \citet{lou2024discrete} made progress by developing score entropy, which extends score matching to discrete spaces and demonstrates substantial performance improvements. While effectively adapting diffusion models to discrete data, these methods have limitations. The discrete nature of the noise limits its expressiveness, making it difficult to capture the nuances of continuous transitions between states. This restricts the model's ability to represent gradual semantic changes or finely adjust individual token features, potentially limiting the benefits of multi-step generation.

\subsection{Continuous Diffusion Models}

Continuous diffusion models map discrete tokens to a continuous vector space using a mapping function, allowing the application of standard continuous diffusion processes. Analog Bits \citep{chen2022analog} uses a binary encoding scheme ($\text{int2bit}: \mathbb{Z} \to { 0, 1 }^{\left\lceil \log_2 V \right\rceil}$) to represent token indices as binary sequences. After the reverse diffusion process, a quantization operation followed by binary decoding ($\text{bit2int}: { 0, 1 }^{\left\lceil \log_2 V \right\rceil} \to \mathbb{Z}$) recovers the token indices. \citet{han2023ssd} proposed a mapping function $\text{logits-generation}: \mathbb{Z} \to \mathbb{R}^V$, which transforms token indices into a probability simplex. \citet{li2022diffusion} proposed Diffusion-LM, where the token sequence $\mathbf{y}$ is first mapped to a random representation $\mathbf{z}_0$ using a word embedding as the mean. After the reverse diffusion process, the generated vectors are rounded back to discrete tokens. \citet{gong2022diffuseq} extended this approach to sequence-to-sequence generation with DiffuSeq, which concatenates the source and target sentences and utilizes an attention mechanism to leverage source information during generation. However, a key limitation of continuous diffusion models is the uniform noise injection applied to all tokens during the forward process. This uniform noise injection hinders the model's ability to effectively leverage contextual information. Ideally, varying noise levels across tokens would allow the model to utilize less noisy tokens as context for restoring more corrupted ones, facilitating better contextual modeling.

\subsection{Improvements over Previous Diffusion Models}
Recent studies have explored various methods to address the limitations discussed above. \citet{han2023ssd} introduced a semi-autoregressive generation strategy that generates fixed-length blocks autoregressively while employing non-autoregressive iterative denoising within each block. \citet{wu2024ar} proposed a hierarchical noise addition method, where noise levels increase monotonically from left to right within a sentence, enabling autoregressive generation. \citet{gong2023diffuseq} presented a hybrid approach that combines standard continuous noise with the probabilistic replacement of tokens with [MASK], integrating discrete and continuous noise. Although these studies have contributed to enhancing the forward diffusion process, their improvements did not fully achieve fine-grained noise at the token level, thus not completely addressing the limitations of both continuous and discrete diffusion models. Also, these approaches typically employ a fixed reverse process that mirrors the forward diffusion process, without considering the varying difficulties in denoising different tokens and the impact of the actual generation context.

\section{Non-simultaeous Continuous Diffusion Models}

\begin{figure*}[t]
    \centering
    \includegraphics[width=1\textwidth]{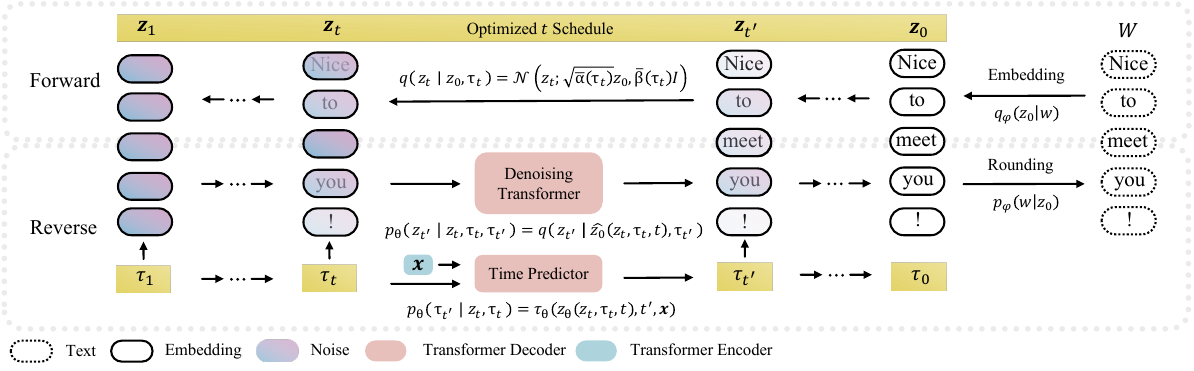}
    \caption{An overview of NeoDiff.}
    \label{fig:overview}
\end{figure*}

To address these limitations, we propose a unified diffusion framework called Non-simultaneous Continuous Diffusion Models (NeoDiff). Figure~\ref{fig:overview} presents an overview of NeoDiff, illustrating its architecture and key components. NeoDiff employs an Encoder-Decoder Transformer architecture\cite{vaswani2017attention}, with the decoder serving as the primary component for denoising, and the encoder provides the embedding of the condition sentence $\textbf{x}$ to a transformer-decoder-based time predictor. In the following sections, we will provide a detailed formulation of NeoDiff and demonstrate how it addresses the limitations of previous approaches.

\subsection{Unified Formulation and Training Objective}

We present a unified framework for diffusion models by introducing two time dimensions: extrinsic time $t$ and intrinsic time $\tau$. The extrinsic time $t$ represents the global diffusion progress of the entire sentence, while the intrinsic time $\tau$ captures the diffusion progress of individual tokens. 

This formulation generalizes existing approaches. We can easily derive discrete diffusion models by modeling $\tau$ as a monotonically increasing random function of $t$, with $\ttau{t} \in \{ 0, 1 \}$, where $\ttau{t} = 0$ and $\ttau{t} = 1$ signify original and fully corrupted tokens, respectively. And continuous diffusion can be obtained by setting $\tau$ as a deterministic function that typically equals $t$ ($\ttau{t} = t$). Furthermore, recent hybrid diffusion models, such as DiffuSeq-V2\cite{gong2023diffuseq}, can also be formalized under this framework by setting $\tau_t = \max(t + \tau_{\text{mask}}(t), 1)$, where $\tau_{\text{mask}}(t) \sim \text{Bernoulli}(\gamma , \bar{\beta}(t))$ and $\gamma$ is the ratio of tokens replaced by [MASK] when $t = 1$.

NeoDiff defines $\tau_t \in [0, 1]$ as a continuous random function of extrinsic time $t \in [0, 1]$, enabling fine-grained control over the diffusion process. We impose boundary conditions $\tau_0 = 0$ and $\tau_1 = 1$ to guarantee token preservation at initialization and complete corruption at termination of the diffusion process.

Let $\mathbf{z} \in \mathbb{R}^d$ denote a token embedding and $\mathbf{z}_t$ its latent representation at time $t$, with initial and final conditions $\mathbf{z}_0 = \mathbf{z}$ and $\mathbf{z}_1 \sim \mathcal{N}(\mathbf{0}, \mathbf{I})$. The forward process defines the joint distribution as:

\begin{align*}
    \qcond{\z{> 0}, \ttau{> 0}} {\z{0}} &:=
    \prod_{t > 0}
    \qcond{\z{t}, \ttau{t}} {\z{0}} \\
    &= \prod_{t > 0}
    \qcond{\z{t}} {\z{0}, \ttau{t}} \q{\ttau{t}},
\end{align*}
where
\begin{equation*}
    \qcond{\z{t}} {\z{0}, \ttau{t}} := \mathcal{N} \left(
        \z{t};
        \sqrt{\bar{\alpha}(\ttau{t})} \z{0},
        \bar{\beta}(\ttau{t}) \mathbf{I}
    \right),
\end{equation*}
and $\bar{\alpha}(\cdot)$ and $\bar{\beta}(\cdot)$ denote noise schedules with their domains scaled to $[0, 1]$.

Given $t' = t - \dt$, the reverse process is defined as
\begin{align*}
    &\p{\z{0:1}, \ttau{0:1}} :=
    \p{\z{1}, \ttau{1}}
    \prod_{t' < 1}
    \pcond{\z{t'}, \ttau{t'}} {\z{t}, \ttau{t}} \\
    &\quad = \p{\z{1}, \ttau{1}}
    \prod_{t' < 1}
    \pcond{\z{t'}} {\z{t}, \ttau{t}, \ttau{t'}}
    \pcond{\ttau{t'}} {\z{t}, \ttau{t}}.
\end{align*}

We further parameterize the distribution of $\z{t'}$ as
\begin{equation*}
    \pcond{\z{t'}} {\z{t}, \ttau{t}, \ttau{t'}} = 
    \qcond{\z{t'}} {\zhat(\z{t}, \ttau{t}, t), \ttau{t'}},
\end{equation*} where $\zhat$ is the model prediction of $\z{0}$. 
    
Following \citet{ho2020denoising} and \citet{li2022diffusion}, we derive NeoDiff's training objective from the variational lower-bound $\lvlb$, and with the simplified $\lz$ and an anchor loss $\lanchor$~\cite{gao-etal-2024-empowering} as a regularization term to avoid collapse of the embedding space, the training objective of Neodiff can be written as  
\begin{align}
    \loss &= \lz + \ltau + \lanchor \\
    &= \mathbb{E}_q\bigg[
        \underbrace{\lVert \zhat(\z{t}, \ttau{t}, t) - \z{0} \rVert^2}_{\lz} \\
        &\quad + \sum_{0 < t' < 1} \underbrace{
            \kl
            {\q{\ttau{t'}}}
            {\pcond{\ttau{t'}} {\z{t}, \ttau{t}}}
        }_{\ltau} \\
        &\quad + \underbrace{-\log \p{y | \zhat(\z{t}, \ttau{t}, t)}}_{\lanchor} 
    \bigg].
\end{align} A detailed derivation can be found in Appendix~\ref{app:detailed_derivation}.

\subsection{Fine-Grained Forward Process Using Poisson Diffusion}

After establishing the unified formulation, we define a fine-grained forward diffusion process through intrinsic time $\tau$. To quantify the diffusion progression within a single token, we introduce a discrete state function $\s{t} \in \{ 0, 1, 2, \cdots, \smax \}$, where uniformly divided states represent distinct levels of the diffusion process from $\s{t} = 0$ (noiseless) to $\s{t} = \smax$ (maximum noise). For an infinitesimal time interval $\dt$, the transition dynamics follow a Poisson process characterized by:
\begin{align*}
    \mathbb{P} [\s{t} = \s{t'} + 1] &= \gamma(t) \dt + o(\dt) \\
    \mathbb{P} [\s{t} = \s{t'}] &= 1 - \gamma(t) \dt + o(\dt),
\end{align*}
where $\gamma(\cdot)$ is a hyperparameter function termed the transition schedule. This formulation yields a tractable distribution for $\s{t}$:
\begin{align*}
    \s{t} &\sim \funcup{Poisson}{\int_0^t \func{\gamma}{t} \dif t} = \funcup{Poisson}{\func{\lambda}{t}}.
\end{align*}
To ensure compatibility with the continuous-time framework of NeoDiff, we normalize the state function to $[0, 1]$ through normalization and clipping: 
\begin{align*}
    \ttau{t} &  =  \funcup{Clip}{\frac{\s{t}}{\smax}, 1} =  \funcup{Clip}{\s{t}', 1},
\end{align*}
where $\funcup{Clip}{\cdot, \cdot}$ denotes the truncation operation to maintain bounded noise levels. 
We choose $\smax$ sufficiently large to achieve fine-grained transitions between noise states, and set $\lambdat = k \smax t$ to maintain $\expect{\s{t}} = \lambdat \propto \smax$. This design ensures that $\ttau{t}$ remains independent of $\smax$ and reduces the process to a homogeneous Poisson process with constant transition schedule $\gamma(t)$.

However, a critical limitation of this basic formulation emerges when examining the coefficient of variation (CV) of the normalized state function $\stpt$:
\begin{equation*}
    \mathrm{CV} = \frac{\sqrt{\var{\s{t}'}}}{\expect{\s{t}'}}
    = \frac{1}{\sqrt{\lambdat}}
    \propto \frac{1}{\sqrt{\smax}},
\end{equation*}
which indicates that as $\smax$ increases, the relative variation between token states diminishes proportionally to $\frac{1}{\sqrt{\smax}}$. Consequently, when $\smax$ becomes sufficiently large, the discreteness of the process is lost as all tokens effectively share nearly identical $\tau$ values, causing NeoDiff to degenerate into a continuous diffusion model.

To address this limitation, we further introduce a variance-controlled rescaling transformation:
\begin{equation}
    \ttau{t} =
    \frac{\funcup{Clip}{\funcup{Round}{
        \frac{\s{t} - \lambdat}{\sqrt{\lambdat}}
        \sigmat + \lambdat
    }, \smax}}{\smax}
    \label{equ:tau}
\end{equation}

Under this transformation, the variables within $\funcup{Clip}{\cdot,\cdot}$ follow a distribution centered at $\lambdat$ with variance $\sigmat$. To ensure that the discrete characteristics of our process remain invariant to the choice of $\smax$, we set $\sigmat = \lambdat$. Since the choice of $\lambdat$ and $\sigmat$ may result in $\tau_{1} \neq 1$, we truncate $\ttau{t}$ to 1 for $t > t_{\mathrm{max}}$, where $t_{\mathrm{max}}$ is a predefined threshold.

\subsection{Context-aware Reverse Process with Time Predictor}

We propose a context-aware reverse process that explicitly models the conditional distribution $\pcond{\ttau{t'}}{\z{t}, \ttau{t}}$, in contrast to previous approaches that simply mirror the forward process by assuming $\pcond{\ttau{t'}}{\z{t}, \ttau{t}} = \q{\ttau{t'}}$. This explicit modeling enables adaptive denoising based on both semantic context and noise states.

\paragraph{Time Predictor Design.} When modeling a known distribution, researchers typically employ reparameterization tricks to model its parameters. However, in our case, the Poisson distribution's sole parameter $\lambdat$ is a deterministic function of $t$ that measures the overall noise progress of the sample and is equivalent to $t$. To obtain the noise progression $\tau$ for each basic token, we directly treat both $\pcond{\ttau{t'}}{\z{t}, \ttau{t}}$ and $\q{\ttau{t'}}$ as standard discrete distributions and learn them using cross-entropy loss, without using reparameterization tricks.

\paragraph{Model Input Design.} 

While $\z{t}$ could serve as an input to $\taupred$, this choice would enable the model to predict noise levels through direct comparison with all other embedding vectors. Such an approach would result in a reverse process that merely retraces the forward process, providing little value for generation quality control. To address this limitation, we propose using the generated sample $\zfpred$ as input to $\taupred$. This design choice increases the modeling complexity of the prediction task while enabling $\taupred$ to serve dual purposes: noise level prediction and semantic quality assessment of the generated output. To provide temporal context, we incorporate $t'=t-\Delta t$ as an additional input, ensuring the model's awareness of the target time distribution. The complete formulation of $\pcond{\ttau{t'}}{\z{t}, \ttau{t}}$ is expressed as $\taupred(\zfpred(\z{t}, \ttau{t}, t), t',\bm{x})$, where $\bm{x}$ represents the conditioning sentence embedding.

\paragraph{Pseudo Label for Training the Time Predictor.} 

The naive approach of using $\ttau{t'}$ as the direct training label for the time predictor can introduce systematic bias in the learning process. While $\ttau{t'}$ is derived from $\zfpred$, this predicted quality measure may not accurately reflect the actual generation quality after the complete denoising process. For example, tokens initially assigned high noise levels might still produce high-quality outputs after denoising, making their initial $\ttau{t'}$ assignment suboptimal. Instead, we propose a pseudo-labeling strategy for training the time predictor. More specifically, we first compute a confidence score for each generated output using the combined loss $\lz + \lanchor$ from the denoised prediction $\zfpred$. To ensure these confidence scores follow a distribution compatible with $\ttau{t'}$, we apply inverse transform sampling. To accomplish this, we compute the normalized rank $r$ for each token's loss within the single sample and map these ranks through the inverse cumulative distribution function (ICDF) of the Poisson distribution: $\func{\tilde{s}}{t} = \func{F^{-1}}{r; \lambdat}$, where $F$ denotes the Poisson cumulative distribution function. The resulting $\func{\tilde{s}}{t}$ values are then transformed via \cref{equ:tau} to obtain the final pseudo labels.

\subsection{Optimized Extrinsic Time Schedule}

The choice of time schedule in diffusion models significantly impacts both generation quality and computational efficiency. While previous works such as \citet{dhariwal2021diffusion,chen2023importancenoiseschedulingdiffusion} focus on optimizing the noise schedule function with fixed extrinsic time steps, we propose to perform direct optimization on the schedule of extrinsic time $t$. Our method builds upon \citet{li-etal-2024-shot-temporal}, who introduced post-training Bayesian optimization to select optimal subsets of time steps for inference acceleration. However, where they treat time steps as discrete variables and optimize for subset selection, we formulate the problem as continuous optimization over the complete time schedule $\{t_1, t_2, \dots, t_K\}$, where $K$ denotes the total number of diffusion steps. This continuous formulation enables more precise calibration through Bayesian optimization, effectively exploring the full space of possible time schedules. We evaluate candidate schedules using a trained model on the validation set via Bayesian optimization, optimizing for the BLEU score as our objective metric. This approach yields task-specific optimal time schedules that further enhances generation quality. The detailed optimization procedure is presented in Appendix~\ref{appendix:time_schedule}.

\section{Experiments}

\subsection{Experimental Setup}

\paragraph{Datasets and Metrics} We evaluate our approach on several NLP tasks, including machine translation (WMT14 En-De~\citep{bojar2014findings}, WMT16 En-Ro~\citep{bojar2016findings}, IWSLT14 De-En~\citep{cettolo2014report}), paraphrasing (QQP), text simplification (Wiki-Auto~\cite{jiang2020neural}), and question generation (Quasar-T~\cite{dhingra2017quasar}). Dataset splits are detailed in Table~\ref{tab:dataset}. We 
use BLEU score~\cite{papineni2002bleu} as the evaluation metric across all tasks, supplemented with SacreBLEU~\cite{post-2018-call} for translation tasks. For comprehensive evaluation, we employ LLM-based evaluation using DeepSeek-V3 685B~\cite{deepseekai2024deepseekv3technicalreport} with specialized prompts, assessing accuracy, fluency, completeness, and task-specific criteria such as creativity for translation and phrasing diversity for paraphrasing. The evaluation process involved providing the LLM with source text, generated text from different models, and specific instructions tailored to each task.  Figure~\ref{fig:prompts} shows the prompt templates used. To rigorously assess the diversity of outputs of the model, We also  included Inter-Sentence Div-4 as \citet{gong2022diffuseq}, which measure diversity at the set-of-outputs-per-source level.

\begin{table}[t] 
    \centering
    \small
    \begin{tabular}{c l c c@{\hspace{2.5pt}}c@{\hspace{2.5pt}}c}
        \toprule
        \textbf{T} & \textbf{Model} & \textbf{$b$} &
        \textbf{IWSLT} & \textbf{WMT14} & \textbf{WMT16} \\
        \midrule
        \multirow{2}{*}{D} 
        & Absorbing & $5$ &
        $28.32^*$ & $21.62^*$ & $30.41^*$ \\        
    
        & Multinomial & $5$ &
        $21.28^*$ & $6.94^*$ & $25.25^*$ \\  
        
        \midrule
        \multirow{6}{*}{C} 
        
        & AR-Diffusion & $1$ &
        $26.78$ & - & - \\
        
        & AR-Diffusion & $10$ &
        $30.64$ & - & - \\
        
        & SeqDiffuSeq & $1$ &
        $28.65^\dag$ & $23.63^\dag$ & $23.98^\dag$ \\
        
        & SeqDiffuSeq & $10$ &
        $30.03^\dag$ & $24.24^\dag$ & $26.17^\dag$ \\

        & Difformer & $1$ &
        $30.94$ & $22.32$ & $30.74$ \\
        
        & Difformer & $10$ &
        $32.09$ & $23.80$ & $30.93$ \\

        \midrule
        
        \ice & \textbf{NeoDiff} & $1$ &
        $32.39^\Uparrow$ & $24.41^\Uparrow$ & $30.87^\Uparrow$ \\
        \ice \multirow{-2}{*}{H} & \textbf{NeoDiff} & $10$ &
        $\textbf{33.14}^\Uparrow$ & $\textbf{25.28}^\Uparrow$ & $\textbf{32.31}^\Uparrow$ \\
        \bottomrule
    \end{tabular}
    \caption{Machine translation BLEU scores for NeoDiff and baseline methods. \textbf{T}: Model type (AR: Autoregressive, D: Discrete, C: Continuous, H: Hybrid). $\Uparrow$: NeoDiff outperforms baselines with beam size $\leq b$; \textbf{bold}: best result. *: Results from \citet{zheng2023reparameterized}; $\dag$: Results from \citet{yuan2023seqdiffuseq}; remaining data reproduced.}
    \label{tab:neodiff_mt} 
\end{table}

\begin{table}[t]
    \centering
    \small
    \begin{tabular}{c l cc@{\hspace{2.5pt}}c@{\hspace{2.5pt}}c}
        \toprule
        \textbf{T} & \textbf{Model} & \textbf{$b$} &
        \textbf{IWSLT} & \textbf{WMT14} & \textbf{WMT16} \\
        \midrule
        \multirow{2}{*}{D} 
         & CMLM & $5$ &
        $29.41^*$ & $23.22^*$ & $31.26^*$ \\
        & CMLM(MBR) &$5$&
        $29.32^*$ & $23.09^*$ & $30.92^*$\\
        \midrule
        \multirow{8}{*}{C} 
         & DiffusionLM & $5$ &
        $26.61^*$ & $15.33^*$ & $27.01^*$ \\  
        
         & DiffusionLM & $50$ &
        $29.11^*$ & $17.41^*$ & $29.39^*$ \\        
        
         & SeqDiffuSeq & $1$ &
        $30.16^\dag$ & $19.16^\dag$ & - \\
        
         & SeqDiffuSeq & $10$ &
        $30.45^\dag$ & $19.76^\dag$ & - \\

         & DiNoiSer & $5$ &
        $31.29^*$ & $24.25^*$ & $30.93^*$ \\
        
         & DiNoiSer & $50$ &
        $31.61^*$ & $24.26^*$ & $31.08^*$ \\

         & Difformer & $1$ &
        $30.06$ & $22.13$ & $30.52$ \\

         & Difformer & $10$ &
        $31.08$ & $23.26$ & $30.75$ \\

        \midrule
         
         \ice & \textbf{NeoDiff} &  $1$ 
         & $31.50^\Uparrow$ &  $24.09$ &  $31.59^\Uparrow$ \\
        
         \ice \multirow{-2}{*}{H} & \textbf{NeoDiff} & $10$ & 
         $\textbf{32.20}^\Uparrow$ & $\textbf{24.64}^\Uparrow$ & $\textbf{32.21}^\Uparrow$ \\
        \bottomrule
    \end{tabular}
    \caption{Comparison on \textbf{SacreBLEU} for machine translation tasks. *: Results from \citet{ye2023dinoiser}; $\dag$: Results from \citet{yuan2023seqdiffuseq}; remaining data are reproduced. $\Uparrow$: NeoDiff outperforms baselines with beam size $\leq b$.}
    \label{tab:neodiff_sacrebleu}
\end{table}

\begin{table}[t]
    \centering
    \small
    \begin{tabular}{cl@{\hspace{1pt}} c@{\hspace{8pt}}c@{\hspace{5pt}}c@{\hspace{5pt}}c@{\hspace{5pt}}}
        \toprule
        \textbf{T} & \textbf{Model} & \textbf{$b$} &
        \textbf{QQP} & \textbf{QT} & \textbf{WA} \\
        \midrule
        \multirow{4}{*}{AR} 
& Transformer & $1$ &
$29.65^{\star}$ & $16.83^{\star}$ & $\textbf{41.68}^{\star}$ \\
& Transformer & $5$ &
$30.83^{\star}$ & $16.45^{\star}$ & $\textbf{43.86}^{\star}$ \\

& GPT2-base FT & - &
$19.80^{\diamond}$ & $7.41^{\diamond}$ & - \\
& GPT2-large FT & - &
$20.59^{\diamond}$ & $11.10^{\diamond}$ & - \\
        \midrule
        \multirow{4}{*}{D} 
         & CMLM & $1$ &
        $24.02$ & - & - \\
         & CMLM & $10$ &
        $26.32$ & - & - \\
    
         & Absorbing & $10$ &
        $23.82^*$ & $17.38^*$ & - \\        
    
         & Multinomial & $10$ &
        $20.70^*$ & $16.96^*$ & - \\  
        
        \midrule
        \multirow{11}{*}{C} 
         & SeqDiffuSeq & $1$ &
        $23.28^\dag$ & $17.20^\dag$ & $37.09^\dag$ \\
        
         & SeqDiffuSeq & $10$ &
        $24.34^\dag$ & $17.46^\dag$ & $37.12^\dag$ \\

         & Difformer & $1$ &
        $28.52$ & $16.03$ & $40.37$ \\
        
         & Difformer & $10$ &
        $30.43$ & $16.66$ & $40.77$ \\
        
        & Meta-DiffuB$^{D_{\theta_1}}$ & - &
        $25.52^\P$ & $18.20^\P$ & $38.77^\P$ \\
        & Meta-DiffuB$^{D_{\theta_2}}$ & - &
        $26.32^\P$ & - & $39.57^\P$ \\
        & Meta-DiffuB$^{D_{\theta_3}}$ & - &
        $22.71^\P$ & - & $24.71^\P$ \\
        & TESS & - &
        $30.20^\ddagger$ & $19.50^\ddagger$ & - \\       
        & TEncDM(BERT) & - &
        $30.20^\circ$ & - & $41.60^\circ$ \\  
        & TEncDM(T5) & - &
        $30.20^\circ$ & - & $41.60^\circ$ \\  
        & TEncDM(RoBERTa) & - &
        $30.00^\circ$ & - & $40.50^\circ$ \\  
        \midrule
    
        H & DiffuSeq-V2 & $1$ &
        $22.10^\S$ & - & - \\
        \midrule
         
         \ice & \textbf{NeoDiff} &  $1$ 
         & $29.47^\Uparrow$ &  $\textbf{20.44}^\Uparrow$ &  41.57 \\
        
         \ice \multirow{-2}{*}{H} & \textbf{NeoDiff} & $10$ & 
         $\textbf{31.32}^\Uparrow$ & $20.03^\Uparrow$ & 41.86 \\
        \bottomrule
    \end{tabular}

\caption{BLEU scores on QQP, QT, and WA(Wiki-Auto). *: Results from \citet{zheng2023reparameterized}; $\dag$: Results from \citet{yuan2023seqdiffuseq}; $\S$: Results from \citet{gong2023diffuseq}; $\star$: Results from \citet{gao-etal-2024-empowering}; $\diamond$: Results from \citet{gong2022diffuseq}; $\P$: Results from \citet{chuang2024metadiffub}. ${D_{\theta_1}}$ = DiffuSeq. ${D_{\theta_2}}$ = SeqDiffuSeq. ${D_{\theta_3}}$ = Dinoiser; $\ddagger$: Results from \citet{karimi-mahabadi-etal-2024-tess}; $\circ$: Results from \citet{shabalin2025tencdmunderstandingpropertiesdiffusion}. Remaining results reproduced. $\Uparrow$: NeoDiff outperforms baselines with beam size $\leq b$.}

    \label{tab:neodiff_other}
\end{table}

\begin{table*}[t]
\small
\centering

\begin{tabular}{@{}>{\raggedright}clcccccc@{}} 
\toprule
 \textbf{Task}& \textbf{Models} & \textbf{b} & \textbf{Semantic Faithfulness} & \textbf{Fluency} & \textbf{Completeness} & \textbf{Phrasing Diversity} \\
\midrule
\multirow{5}{*}{{\textbf{QQP}}} 
& CMLM & 10 & 72.86 & 81.99 & 75.60 & 55.86 \\
& Transformer & 1 & 83.64 & 92.56 & 84.96 & \textbf{57.19} \\
& Transformer & 5 & 83.70 & \textbf{94.73} & 86.05 & 54.52 \\
& Transformer & 10 & 83.93 & 94.64 & 86.02 & 54.55 \\

& NeoDiff & 10 & \textbf{87.42} & 91.87 & \textbf{88.79} & 45.83 \\[0.2ex]
\midrule[0.5pt]
 & \textbf{Models} & \textbf{b} & \textbf{Accuracy} & \textbf{Fluency} & \textbf{Completeness} & \textbf{Creativity} \\
 \midrule[0.4pt]
\multirow{3}{*}{{{\textbf{WMT14}}}}
& Difformer & 10 & 79.72 & 80.31 & 85.24 & 75.12 \\
& Transformer & 5 & \textbf{85.66} & \textbf{86.35} & \textbf{90.81} & \textbf{80.07} \\
& NeoDiff & 10 & 80.30 & 80.81 & 85.61 & 76.20 

\\[0.2ex]
\bottomrule
\end{tabular}
\caption{LLM evaluation of text generation tasks using DeepSeek-v3 685B. We evaluate Paraphrasing (QQP Dataset) and Machine Translation (WMT14 En-De Dataset). (1) We access QQP on Semantic Faithfulness, Fluency, Completeness, and Phrasing Diversity. (2) We access WMT14 En-De on Accuracy, Fluency, Completeness, and Creativity. Detailed prompts are provided in Fig~\ref{fig:prompts}.}
\label{tab:llm_evaluation_combined}

\end{table*}

\paragraph{Baselines} We compared NeoDiff against several strong baselines across 
multiple diffusion model categories. For discrete diffusion models, we included Absorbing Diffusion~\cite{austin2021structured}, Multinomial Diffusion~\cite{NEURIPS2021_67d96d45}, and CMLM~\cite{ghazvininejad-etal-2019-mask}. For continuous diffusion models, we benchmarked against DiffusionLM~\cite{li2022diffusion}, Difformer~\cite{gao-etal-2024-empowering}, SeqDiffuSeq~\cite{yuan2023seqdiffuseq}, AR-Diffusion~\cite{wu2024ar}, DiNoiSer~\cite{ye2024dinoiserdiffusedconditionalsequence}, Meta-DiffuB~\cite{chuang2024metadiffub}, TESS~\cite{karimi-mahabadi-etal-2024-tess} and TEncDM~\cite{shabalin2025tencdmunderstandingpropertiesdiffusion}. For hybrid approaches, we compared with DiffuSeq-V2~\cite{gong2023diffuseq}. We also included Transformer and fine-tuned GPT2 models as autoregressive baselines.

\paragraph{Implementation Details}

We set the maximum noise state $\smax$ to 100 for all tasks and datasets, incorporating self-conditioning \cite{chen2022analog} and noise rescaling with $\dgsmax=0.2$ \cite{gao-etal-2024-empowering}. We used byte pair encoding \cite{sennrich-etal-2016-neural} without knowledge distillation \cite{kim2016sequence} to evaluate under challenging conditions. During decoding, we employed 2D parallel decoding \cite{gao-etal-2024-empowering} and selected the best candidate sentence using the minimum Bayes risk (MBR) method \cite{kumar-byrne-2004-minimum} based on the BLEU score. We also used post-training Bayesian optimization \cite{li-etal-2024-shot-temporal} to calibrate the extrinsic time schedule, limiting the optimization to 100 rounds for all tasks. Details of the experimental settings are provided in Appendix~\ref{app:experimental_settings}.

\subsection{Results}

Our experimental evaluation demonstrates NeoDiff's effectiveness across multiple generation tasks. On machine translation benchmarks (Table~\ref{tab:neodiff_mt} and~\ref{tab:neodiff_sacrebleu}), NeoDiff consistently outperforms existing non-autoregressive diffusion-based, iteration-based, and autoregressive diffusion approaches. As shown in Table~\ref{tab:neodiff_other}, these improvements extend beyond translation to diverse generation tasks. Unlike baselines such as AR-Diffusion that rely heavily on MBR and show performance drops with single samples ($b=1$), NeoDiff maintains robust performance even in this constrained setting. NeoDiff also demonstrates strong performance in LLM-based evaluations (Table~\ref{tab:llm_evaluation_combined}, prompts in Figure~\ref{fig:prompts}). Notably, on the QQP task (Table~\ref{tab:llm_evaluation_combined}), NeoDiff achieves superior scores in semantic faithfulness and completeness. For the WMT14 task, NeoDiff achieves performance comparable to Difformer across multiple aspects. NeoDiff also demonstrates strong inter-sentence diversity (Inter-Sentence Div-4) when generating multiple candidates. Detailed comparisons against AR model on QQP dataset can be found in Appendix~\ref{app:additional_qqp_diversity_analysis} (Table~\ref{tab:appendix_qqp_diversity_comparison}). Our results show that NeoDiff can balance the quality-diversity trade-off more effectively than autoregressive models like Transformer as the output space scales (i.e., with increasing $b$), a characteristic also observed in \citet{gong2022diffuseq}. The Bayesian Optimization component introduces a manageable overhead (Appendix~\ref{app:bayesian}, approximately 6\% of training time on WMT14). Also, NeoDiff's inference speed and memory usage are competitive with similar models (Appendix~\ref{sec:runtime}).

\begin{table}[htb]
    \small
    \centering

    \begin{tabular}{
        @{}C{1.5em} @{}
        @{}C{7em}@{}
        @{}C{6.8em}@{}
        @{}C{6em}@{}
        @{}C{3em}@{}
    }
        \toprule
        \textbf{\#} & \textbf{Poisson Diffusion Process} & \textbf{Time Predictor} & \textbf{Optimized $t$ Schedule} &\textbf{BLEU} \\
        \midrule
        \textbf{Base} &&&& $32.09$ \\
        \textbf{+P} & \checkmark &&& $32.75$ \\
        \textbf{+PT} & \checkmark &\checkmark&&  $32.97$ \\
        \textbf{Full} & \checkmark &\checkmark&\checkmark  & $\textbf{33.14}$ \\
        \bottomrule
    \end{tabular}
    \caption{Ablation study on the impact of proposed components on IWSLT14 De-En dataset with a $b=10$. 
    }
    \label{tab:neodiff_ablation}
\end{table}

\begin{table}[t]
\small
\centering
\begin{tabular}{@{}c@{\hspace{6pt}}p{7.2cm}@{}}
\toprule
\textbf{A} & \textbf{Src:} das zeigt die enorm große rolle , die ein meeresschutzgebiet spielen kann. \\
& \textbf{Ref:} and hence , the enormous role that a marine protected area can play. \\
& \textbf{Base:} and this shows the enormously big role that a \textbf{\color{red}{area can play with a sea protected}} \\
& \textbf{+P:} so this shows the enormously big role that a \textbf{\dg{marine protected area can play}} \\
\midrule
\textbf{B} & \textbf{Src:} er ist ganz glücklich darüber, weil er sie getäuscht hat. \\
& \textbf{Ref:} he'll be very happy because he's deceived you. \\
& \textbf{+P:} he's very happy about it because he \textbf{\red{decaked}} her. \\
& \textbf{+PT:} he's very happy about it because he \textbf{\dg{deceived}} her. \\
\midrule
\textbf{C} & \textbf{Src:} die korrelation ist also gering . \\
& \textbf{Ref:} so the correlation is low . \\
& \textbf{+PT:} so \textbf{\red{it's a small of the correlation}}. \\
& \textbf{Full:} so \textbf{\dg{the correlation is small}}. \\
\bottomrule
\end{tabular}
\caption{Example outputs illustrating three key mechanisms of NeoDiff: (A) improved phrase-level coherence with Poisson process, (B) enhanced token-level refinements with time predictor, and (C) better sentence-level organization with optimized schedule.}
\label{tab:cases}
\end{table}
\subsection{Analysis} Our ablation studies (Table~\ref{tab:neodiff_ablation}) demonstrate clear improvements from each component, with the full model achieving a substantial +1.05 BLEU improvement over the baseline. 
We further analyze each component's impact on generation quality:
\paragraph{Poisson Process for Multi-token Coherence} The Poisson diffusion process enables more fine-grained control over multiple tokens by precise inter-token coordination. This advantage yields a substantial performance gain over standard continuous diffusion ($\tau_t = t$). As evidenced in Table~\ref{tab:cases}A, this improved control manifests itself in better phrase-level coherence.
\paragraph{Time Predictor for Guided Denoising} By leveraging information from less-noised tokens to guide the denoising trajectory of noisier ones, the time predictor enhances the model's ability of more contextually informed token generation. Table~\ref{tab:cases}B demonstrates this through more natural word selections and verb choices that better preserve the original meaning.
\paragraph{Optimized Schedule for Global Coherence} The optimized extrinsic time schedule enables dynamic adjustments to the diffusion trajectory, 
facilitating escape from sub-optimal samples where sequence order or overall structure significantly deviates from the target distribution. This global refinement allows for more substantial rewriting when needed, as demonstrated in Table~\ref{tab:cases}C where entire phrases are better reorganized.

Additional examples demonstrating the impact of these components are provided in Table~\ref{tab:appendix_poisson},~\ref{tab:appendix_timepredictor}, and~\ref{tab:appendix_schedule}. In Appendix~\ref{sec:appendix_translation_examples}, we track the step-wise generation processes, demonstrating superior convergence speed and accuracy for NeoDiff compared to continuous diffusion baselines. We also compared NeoDiff against continuous diffusion baselines on token-level controlled generation, demonstrating its unique ability to perform targeted modifications while maintaining semantic consistency across translations (Appendix~\ref{appendix:token_examples}).

\section{Conclusion}

In this work, we introduce Non-simultaeous Continuous Diffusion Models (NeoDiff), a novel diffusion-based text generation framework that unifies discrete and continuous diffusion models. NeoDiff generalizes the time variable, incorporates the Poisson diffusion process, adaptively modulates the reverse process based on semantic context, and uses an optimized extrinsic time schedule for inference. This unified framework enables fine-grained control and achieves superior performance across diverse natural language processing tasks. Our extensive experiments demonstrate the effectiveness of this unified framework, opening up new avenues for advancing diffusion-based text generation.

\section*{Limitations}

While NeoDiff demonstrates strong performance across various Seq2Seq-based conditional generation tasks (e.g., machine translation, paraphrasing, text simplification, and question generation), we note some implementation considerations. The post-training optimization of extrinsic time schedules requires additional sampling iterations, though this overhead is negligible compared to the training time. The time predictor introduces a modest parameter increase to the backbone network.

\section*{Acknowledgements}

This research was supported by the National Natural Science Foundation
of China (Grant No. 62276245).

\bibliography{custom}

\begin{thebibliography}{44}
\providecommand{\natexlab}[1]{#1}

\bibitem[{Austin et~al.(2021)Austin, Johnson, Ho, Tarlow, and van~den Berg}]{austin2021structured}
Jacob Austin, Daniel~D Johnson, Jonathan Ho, Daniel Tarlow, and Rianne van~den Berg. 2021.
\newblock Structured denoising diffusion models in discrete state-spaces.
\newblock \emph{Advances in Neural Information Processing Systems}, 34:17981--17993.

\bibitem[{Bojar et~al.(2014)Bojar, Buck, Federmann, Haddow, Koehn, Leveling, Monz, Pecina, Post, Saint-Amand et~al.}]{bojar2014findings}
Ond{\v{r}}ej Bojar, Christian Buck, Christian Federmann, Barry Haddow, Philipp Koehn, Johannes Leveling, Christof Monz, Pavel Pecina, Matt Post, Herve Saint-Amand, et~al. 2014.
\newblock Findings of the 2014 workshop on statistical machine translation.
\newblock In \emph{Proceedings of the ninth workshop on statistical machine translation}, pages 12--58.

\bibitem[{Bojar et~al.(2016)Bojar, Chatterjee, Federmann, Graham, Haddow, Huck, Yepes, Koehn, Logacheva, Monz et~al.}]{bojar2016findings}
Ondrej Bojar, Rajen Chatterjee, Christian Federmann, Yvette Graham, Barry Haddow, Matthias Huck, Antonio~Jimeno Yepes, Philipp Koehn, Varvara Logacheva, Christof Monz, et~al. 2016.
\newblock Findings of the 2016 conference on machine translation (wmt16).
\newblock In \emph{First conference on machine translation}, pages 131--198. Association for Computational Linguistics.

\bibitem[{Brochu et~al.(2011)Brochu, Hoffman, and de~Freitas}]{brochu2011portfolio}
Eric Brochu, Matthew~W. Hoffman, and Nando de~Freitas. 2011.
\newblock \href {https://arxiv.org/abs/1009.5419} {Portfolio allocation for bayesian optimization}.
\newblock \emph{Preprint}, arXiv:1009.5419.

\bibitem[{Cettolo et~al.(2014)Cettolo, Niehues, St{\"u}ker, Bentivogli, and Federico}]{cettolo2014report}
Mauro Cettolo, Jan Niehues, Sebastian St{\"u}ker, Luisa Bentivogli, and Marcello Federico. 2014.
\newblock Report on the 11th iwslt evaluation campaign.
\newblock In \emph{Proceedings of the 11th International Workshop on Spoken Language Translation: Evaluation Campaign}, pages 2--17.

\bibitem[{Chen et~al.(2023)Chen, Zhang, Li, Smola, and Yang}]{chen-etal-2023-cheaper}
Jiaao Chen, Aston Zhang, Mu~Li, Alex Smola, and Diyi Yang. 2023.
\newblock \href {https://doi.org/10.18653/v1/2023.emnlp-main.289} {A cheaper and better diffusion language model with soft-masked noise}.
\newblock In \emph{Proceedings of the 2023 Conference on Empirical Methods in Natural Language Processing}, pages 4765--4775, Singapore. Association for Computational Linguistics.

\bibitem[{Chen et~al.(2020)Chen, Zhang, Zen, Weiss, Norouzi, and Chan}]{chen2020wavegrad}
Nanxin Chen, Yu~Zhang, Heiga Zen, Ron~J Weiss, Mohammad Norouzi, and William Chan. 2020.
\newblock Wavegrad: Estimating gradients for waveform generation.
\newblock In \emph{International Conference on Learning Representations}.

\bibitem[{Chen(2023)}]{chen2023importancenoiseschedulingdiffusion}
Ting Chen. 2023.
\newblock \href {https://arxiv.org/abs/2301.10972} {On the importance of noise scheduling for diffusion models}.
\newblock \emph{Preprint}, arXiv:2301.10972.

\bibitem[{Chen et~al.(2022)Chen, ZHANG, and Hinton}]{chen2022analog}
Ting Chen, Ruixiang ZHANG, and Geoffrey Hinton. 2022.
\newblock Analog bits: Generating discrete data using diffusion models with self-conditioning.
\newblock In \emph{The Eleventh International Conference on Learning Representations}.

\bibitem[{Chuang et~al.(2024)Chuang, Hsu, Lin, Gu, Li, Chang, and yi~Lee}]{chuang2024metadiffub}
Yunyen Chuang, Hung-Min Hsu, Kevin Lin, Chen-Sheng Gu, Ling~Zhen Li, Ray-I Chang, and Hung yi~Lee. 2024.
\newblock \href {https://openreview.net/forum?id=NTWXVvIXJM} {Meta-diffu\$b\$: A contextualized sequence-to-sequence text diffusion model with meta-exploration}.
\newblock In \emph{The Thirty-eighth Annual Conference on Neural Information Processing Systems}.

\bibitem[{DeepSeek-AI(2024)}]{deepseekai2024deepseekv3technicalreport}
DeepSeek-AI. 2024.
\newblock \href {https://arxiv.org/abs/2412.19437} {Deepseek-v3 technical report}.
\newblock \emph{Preprint}, arXiv:2412.19437.

\bibitem[{Dhariwal and Nichol(2021)}]{dhariwal2021diffusion}
Prafulla Dhariwal and Alexander Nichol. 2021.
\newblock Diffusion models beat gans on image synthesis.
\newblock \emph{Advances in Neural Information Processing Systems}, 34:8780--8794.

\bibitem[{Dhingra et~al.(2017)Dhingra, Mazaitis, and Cohen}]{dhingra2017quasar}
Bhuwan Dhingra, Kathryn Mazaitis, and William~W. Cohen. 2017.
\newblock \href {https://arxiv.org/abs/1707.03904} {Quasar: Datasets for question answering by search and reading}.
\newblock \emph{Preprint}, arXiv:1707.03904.

\bibitem[{Gao et~al.(2024)Gao, Guo, Tan, Zhu, Zhang, Bian, and Xu}]{gao-etal-2024-empowering}
Zhujin Gao, Junliang Guo, Xu~Tan, Yongxin Zhu, Fang Zhang, Jiang Bian, and Linli Xu. 2024.
\newblock \href {https://doi.org/10.18653/v1/2024.naacl-long.261} {Empowering diffusion models on the embedding space for text generation}.
\newblock In \emph{Proceedings of the 2024 Conference of the North American Chapter of the Association for Computational Linguistics: Human Language Technologies (Volume 1: Long Papers)}, pages 4664--4683, Mexico City, Mexico. Association for Computational Linguistics.

\bibitem[{Ghazvininejad et~al.(2019)Ghazvininejad, Levy, Liu, and Zettlemoyer}]{ghazvininejad-etal-2019-mask}
Marjan Ghazvininejad, Omer Levy, Yinhan Liu, and Luke Zettlemoyer. 2019.
\newblock \href {https://doi.org/10.18653/v1/D19-1633} {Mask-predict: Parallel decoding of conditional masked language models}.
\newblock In \emph{Proceedings of the 2019 Conference on Empirical Methods in Natural Language Processing and the 9th International Joint Conference on Natural Language Processing (EMNLP-IJCNLP)}, pages 6112--6121, Hong Kong, China. Association for Computational Linguistics.

\bibitem[{Gong et~al.(2022)Gong, Li, Feng, Wu, and Kong}]{gong2022diffuseq}
Shansan Gong, Mukai Li, Jiangtao Feng, Zhiyong Wu, and Lingpeng Kong. 2022.
\newblock Diffuseq: Sequence to sequence text generation with diffusion models.
\newblock In \emph{The Eleventh International Conference on Learning Representations}.

\bibitem[{Gong et~al.(2023)Gong, Li, Feng, Wu, and Kong}]{gong2023diffuseq}
Shansan Gong, Mukai Li, Jiangtao Feng, Zhiyong Wu, and Lingpeng Kong. 2023.
\newblock Diffuseq-v2: Bridging discrete and continuous text spaces for accelerated seq2seq diffusion models.
\newblock In \emph{Findings of the Association for Computational Linguistics: EMNLP 2023}, pages 9868--9875.

\bibitem[{Han et~al.(2023)Han, Kumar, and Tsvetkov}]{han2023ssd}
Xiaochuang Han, Sachin Kumar, and Yulia Tsvetkov. 2023.
\newblock Ssd-lm: Semi-autoregressive simplex-based diffusion language model for text generation and modular control.
\newblock In \emph{Proceedings of the 61st Annual Meeting of the Association for Computational Linguistics (Volume 1: Long Papers)}, pages 11575--11596.

\bibitem[{Ho et~al.(2020)Ho, Jain, and Abbeel}]{ho2020denoising}
Jonathan Ho, Ajay Jain, and Pieter Abbeel. 2020.
\newblock Denoising diffusion probabilistic models.
\newblock \emph{Advances in Neural Information Processing Systems}, 33:6840--6851.

\bibitem[{Ho and Salimans(2021)}]{ho2021classifier}
Jonathan Ho and Tim Salimans. 2021.
\newblock Classifier-free diffusion guidance.
\newblock In \emph{NeurIPS 2021 Workshop on Deep Generative Models and Downstream Applications}.

\bibitem[{Hoogeboom et~al.(2021{\natexlab{a}})Hoogeboom, Nielsen, Jaini, Forr{\'e}, and Welling}]{hoogeboom2021argmax}
Emiel Hoogeboom, Didrik Nielsen, Priyank Jaini, Patrick Forr{\'e}, and Max Welling. 2021{\natexlab{a}}.
\newblock Argmax flows and multinomial diffusion: Learning categorical distributions.
\newblock \emph{Advances in Neural Information Processing Systems}, 34:12454--12465.

\bibitem[{Hoogeboom et~al.(2021{\natexlab{b}})Hoogeboom, Nielsen, Jaini, Forr\'{e}, and Welling}]{NEURIPS2021_67d96d45}
Emiel Hoogeboom, Didrik Nielsen, Priyank Jaini, Patrick Forr\'{e}, and Max Welling. 2021{\natexlab{b}}.
\newblock \href {https://proceedings.neurips.cc/paper_files/paper/2021/file/67d96d458abdef21792e6d8e590244e7-Paper.pdf} {Argmax flows and multinomial diffusion: Learning categorical distributions}.
\newblock In \emph{Advances in Neural Information Processing Systems}, volume~34, pages 12454--12465. Curran Associates, Inc.

\bibitem[{Jiang et~al.(2020)Jiang, Maddela, Lan, Zhong, and Xu}]{jiang2020neural}
Chao Jiang, Mounica Maddela, Wuwei Lan, Yang Zhong, and Wei Xu. 2020.
\newblock Neural crf model for sentence alignment in text simplification.
\newblock In \emph{Proceedings of the 58th Annual Meeting of the Association for Computational Linguistics}, pages 7943--7960.

\bibitem[{Karimi~Mahabadi et~al.(2024)Karimi~Mahabadi, Ivison, Tae, Henderson, Beltagy, Peters, and Cohan}]{karimi-mahabadi-etal-2024-tess}
Rabeeh Karimi~Mahabadi, Hamish Ivison, Jaesung Tae, James Henderson, Iz~Beltagy, Matthew Peters, and Arman Cohan. 2024.
\newblock \href {https://aclanthology.org/2024.eacl-long.144/} {{TESS}: Text-to-text self-conditioned simplex diffusion}.
\newblock In \emph{Proceedings of the 18th Conference of the European Chapter of the Association for Computational Linguistics (Volume 1: Long Papers)}, pages 2347--2361, St. Julian{'}s, Malta. Association for Computational Linguistics.

\bibitem[{Kim and Rush(2016)}]{kim2016sequence}
Yoon Kim and Alexander~M Rush. 2016.
\newblock Sequence-level knowledge distillation.
\newblock In \emph{Proceedings of the 2016 Conference on Empirical Methods in Natural Language Processing}, pages 1317--1327.

\bibitem[{Kong et~al.(2020)Kong, Ping, Huang, Zhao, and Catanzaro}]{kong2020diffwave}
Zhifeng Kong, Wei Ping, Jiaji Huang, Kexin Zhao, and Bryan Catanzaro. 2020.
\newblock Diffwave: A versatile diffusion model for audio synthesis.
\newblock In \emph{International Conference on Learning Representations}.

\bibitem[{Kumar and Byrne(2004)}]{kumar-byrne-2004-minimum}
Shankar Kumar and William Byrne. 2004.
\newblock \href {https://aclanthology.org/N04-1022} {Minimum {B}ayes-risk decoding for statistical machine translation}.
\newblock In \emph{Proceedings of the Human Language Technology Conference of the North {A}merican Chapter of the Association for Computational Linguistics: {HLT}-{NAACL} 2004}, pages 169--176, Boston, Massachusetts, USA. Association for Computational Linguistics.

\bibitem[{Li et~al.(2024)Li, Gao, Zhu, Yin, Cao, Jiang, and Xu}]{li-etal-2024-shot-temporal}
Bocheng Li, Zhujin Gao, Yongxin Zhu, Kun Yin, Haoyu Cao, Deqiang Jiang, and Linli Xu. 2024.
\newblock \href {https://aclanthology.org/2024.lrec-main.637} {Few-shot temporal pruning accelerates diffusion models for text generation}.
\newblock In \emph{Proceedings of the 2024 Joint International Conference on Computational Linguistics, Language Resources and Evaluation (LREC-COLING 2024)}, pages 7259--7269, Torino, Italia. ELRA and ICCL.

\bibitem[{Li et~al.(2022)Li, Thickstun, Gulrajani, Liang, and Hashimoto}]{li2022diffusion}
Xiang Li, John Thickstun, Ishaan Gulrajani, Percy~S Liang, and Tatsunori~B Hashimoto. 2022.
\newblock Diffusion-lm improves controllable text generation.
\newblock \emph{Advances in Neural Information Processing Systems}, 35:4328--4343.

\bibitem[{Liu and Nocedal(1989)}]{Liu1989OnTL}
Dong~C. Liu and Jorge Nocedal. 1989.
\newblock On the limited memory bfgs method for large scale optimization.
\newblock \emph{Mathematical Programming}, 45:503--528.

\bibitem[{Lou et~al.(2024)Lou, Meng, and Ermon}]{lou2024discrete}
Aaron Lou, Chenlin Meng, and Stefano Ermon. 2024.
\newblock Discrete diffusion modeling by estimating the ratios of the data distribution.
\newblock \emph{arXiv preprint arXiv:2310.16834}.

\bibitem[{Nichol and Dhariwal(2021)}]{nichol2021improved}
Alexander~Quinn Nichol and Prafulla Dhariwal. 2021.
\newblock Improved denoising diffusion probabilistic models.
\newblock In \emph{International Conference on Machine Learning}, pages 8162--8171. PMLR.

\bibitem[{Ott et~al.(2019)Ott, Edunov, Baevski, Fan, Gross, Ng, Grangier, and Auli}]{ott2019fairseq}
Myle Ott, Sergey Edunov, Alexei Baevski, Angela Fan, Sam Gross, Nathan Ng, David Grangier, and Michael Auli. 2019.
\newblock fairseq: A fast, extensible toolkit for sequence modeling.
\newblock In \emph{Proceedings of NAACL-HLT 2019: Demonstrations}.

\bibitem[{Papineni et~al.(2002)Papineni, Roukos, Ward, and Zhu}]{papineni2002bleu}
Kishore Papineni, Salim Roukos, Todd Ward, and Wei-Jing Zhu. 2002.
\newblock Bleu: a method for automatic evaluation of machine translation.
\newblock In \emph{Proceedings of the 40th annual meeting of the Association for Computational Linguistics}, pages 311--318.

\bibitem[{Post(2018)}]{post-2018-call}
Matt Post. 2018.
\newblock \href {https://www.aclweb.org/anthology/W18-6319} {A call for clarity in reporting {BLEU} scores}.
\newblock In \emph{Proceedings of the Third Conference on Machine Translation: Research Papers}, pages 186--191, Belgium, Brussels. Association for Computational Linguistics.

\bibitem[{Rombach et~al.(2022)Rombach, Blattmann, Lorenz, Esser, and Ommer}]{rombach2022high}
Robin Rombach, Andreas Blattmann, Dominik Lorenz, Patrick Esser, and Bj{\"o}rn Ommer. 2022.
\newblock High-resolution image synthesis with latent diffusion models.
\newblock In \emph{Proceedings of the IEEE/CVF Conference on Computer Vision and Pattern Recognition}, pages 10684--10695.

\bibitem[{Sennrich et~al.(2016)Sennrich, Haddow, and Birch}]{sennrich-etal-2016-neural}
Rico Sennrich, Barry Haddow, and Alexandra Birch. 2016.
\newblock \href {https://doi.org/10.18653/v1/P16-1162} {Neural machine translation of rare words with subword units}.
\newblock In \emph{Proceedings of the 54th Annual Meeting of the Association for Computational Linguistics (Volume 1: Long Papers)}, pages 1715--1725, Berlin, Germany. Association for Computational Linguistics.

\bibitem[{Shabalin et~al.(2025)Shabalin, Meshchaninov, Chimbulatov, Lapikov, Kim, Bartosh, Molchanov, Markov, and Vetrov}]{shabalin2025tencdmunderstandingpropertiesdiffusion}
Alexander Shabalin, Viacheslav Meshchaninov, Egor Chimbulatov, Vladislav Lapikov, Roman Kim, Grigory Bartosh, Dmitry Molchanov, Sergey Markov, and Dmitry Vetrov. 2025.
\newblock \href {https://arxiv.org/abs/2402.19097} {Tencdm: Understanding the properties of the diffusion model in the space of language model encodings}.
\newblock \emph{Preprint}, arXiv:2402.19097.

\bibitem[{Vaswani et~al.(2017)Vaswani, Shazeer, Parmar, Uszkoreit, Jones, Gomez, Kaiser, and Polosukhin}]{vaswani2017attention}
Ashish Vaswani, Noam Shazeer, Niki Parmar, Jakob Uszkoreit, Llion Jones, Aidan~N Gomez, {\L}ukasz Kaiser, and Illia Polosukhin. 2017.
\newblock Attention is all you need.
\newblock \emph{Advances in neural information processing systems}, 30.

\bibitem[{Wu et~al.(2024)Wu, Fan, Liu, Zheng, Gong, Jiao, Li, Guo, Duan, Chen et~al.}]{wu2024ar}
Tong Wu, Zhihao Fan, Xiao Liu, Hai-Tao Zheng, Yeyun Gong, Jian Jiao, Juntao Li, Jian Guo, Nan Duan, Weizhu Chen, et~al. 2024.
\newblock Ar-diffusion: Auto-regressive diffusion model for text generation.
\newblock \emph{Advances in Neural Information Processing Systems}, 36.

\bibitem[{Ye et~al.(2023)Ye, Zheng, Bao, Qian, and Wang}]{ye2023dinoiser}
Jiasheng Ye, Zaixiang Zheng, Yu~Bao, Lihua Qian, and Mingxuan Wang. 2023.
\newblock \href {https://arxiv.org/abs/2302.10025} {Dinoiser: Diffused conditional sequence learning by manipulating noises}.
\newblock \emph{Preprint}, arXiv:2302.10025.

\bibitem[{Ye et~al.(2024)Ye, Zheng, Bao, Qian, and Wang}]{ye2024dinoiserdiffusedconditionalsequence}
Jiasheng Ye, Zaixiang Zheng, Yu~Bao, Lihua Qian, and Mingxuan Wang. 2024.
\newblock \href {https://arxiv.org/abs/2302.10025} {Dinoiser: Diffused conditional sequence learning by manipulating noises}.
\newblock \emph{Preprint}, arXiv:2302.10025.

\bibitem[{Yuan et~al.(2024)Yuan, Yuan, Tan, Huang, and Huang}]{yuan2023seqdiffuseq}
Hongyi Yuan, Zheng Yuan, Chuanqi Tan, Fei Huang, and Songfang Huang. 2024.
\newblock \href {https://doi.org/10.18653/v1/2024.naacl-long.2} {Text diffusion model with encoder-decoder transformers for sequence-to-sequence generation}.
\newblock In \emph{Proceedings of the 2024 Conference of the North American Chapter of the Association for Computational Linguistics: Human Language Technologies (Volume 1: Long Papers)}, pages 22--39, Mexico City, Mexico. Association for Computational Linguistics.

\bibitem[{Zheng et~al.(2023)Zheng, Yuan, Yu, and Kong}]{zheng2023reparameterized}
Lin Zheng, Jianbo Yuan, Lei Yu, and Lingpeng Kong. 2023.
\newblock A reparameterized discrete diffusion model for text generation.
\newblock \emph{arXiv preprint arXiv:2302.05737}.

\end{thebibliography}

\appendix

\section{Detailed Derivation of the Training Objective of NeoDiff}
\label{app:detailed_derivation}
Let $\boldsymbol{z}$ represent a token embedding and $\z{t}$ its latent representation at time $t$, with $\z{0} = \boldsymbol{z}$ and $\z{1} \sim \mathcal{N}(\boldsymbol{0}, \mathbf{I})$. The joint distribution of the forward process is then given by:

\begin{align}
    \qcond{\z{> 0}, \ttau{> 0}} {\z{0}} &:=
    \prod_{t > 0}
    \qcond{\z{t}, \ttau{t}} {\z{0}} \\
    &= \prod_{t > 0}
    \qcond{\z{t}} {\z{0}, \ttau{t}} \q{\ttau{t}},
\end{align}
where
\begin{equation*}
    \qcond{\z{t}} {\z{0}, \ttau{t}} := \mathcal{N} \left(
        \z{t};
        \sqrt{\bar{\alpha}(\ttau{t})} \z{0},
        \bar{\beta}(\ttau{t}) \mathbf{I}
    \right),
\end{equation*}
and $\bar{\alpha}(\cdot)$ and $\bar{\beta}(\cdot)$ denote noise schedules with their domains scaled to $[0, 1]$.

Given $t' = t - \dt$, the reverse process is defined as
\begin{align}
    &\p{\z{0:1}, \ttau{0:1}} :=
    \p{\z{1}, \ttau{1}}
    \prod_{t' < 1}
    \pcond{\z{t'}, \ttau{t'}} {\z{t}, \ttau{t}} \\
    &\quad = \p{\z{1}, \ttau{1}}
    \prod_{t' < 1}
    \pcond{\z{t'}} {\z{t}, \ttau{t}, \ttau{t'}}
    \pcond{\ttau{t'}} {\z{t}, \ttau{t}}.
\end{align}

We further parameterize the distribution of $\z{t'}$ as
\begin{equation*}
    \pcond{\z{t'}} {\z{t}, \ttau{t}, \ttau{t'}} = 
    \qcond{\z{t'}} {\zhat(\z{t}, \ttau{t}, t), \ttau{t'}},
\end{equation*} where $\zhat$ is the model prediction of $\z{0}$. 
    
Following \citet{ho2020denoising}, the training objective is derived from the variational lower-bound
\begin{align}
    \lvlb &=
    \mathbb{E}_q \left[
        -\log \frac
        {\p{\z{0:1}, \ttau{0:1}}}
        {\qcond{\z{> 0}, \ttau{> 0}} {\z{0}}}
    \right] \\
    &= \mathbb{E}_q \Biggl[
        -\log \frac
        {\p{\z{1}, \ttau{1}}}
        {\qcond{\z{1}, \ttau{1}} {\z{0}}} \\
        &\quad +\sum_{0 < t' < 1} -\log \frac {
            \qcond{\z{t'}} {\zhat(\z{t}, \ttau{t}, t), \ttau{t'}}
        } {\qcond{\z{t'}} {\z{0}, \ttau{t'}}} \\
        &\quad +\sum_{0 < t' < 1} -\log \frac
        {\pcond{\ttau{t'}} {\z{t}, \ttau{t}}}
        {\q{\ttau{t'}}} \\
        &\quad -\log \pcond
        {\z{0}, \ttau{0}}
        {\z{\dt}, \ttau{\dt}}
    \Biggr] \\
    &= \mathbb{E}_q \Biggl[
        \underbrace{
            \kl
            {\qcond{\z{1}, \ttau{1}} {\z{0}}}
            {\p{\z{1}, \ttau{1}}}
        }_{\loss_1} \\
        &\quad +\sum_{0 < t' < 1} \underbrace{
            \kl
            {\qcond{\z{t'}} {\z{0}, \ttau{t'}}} {
                \qcond{\z{t'}} {\zhat, \ttau{t'}}
            }
        }_{\lz} \\
        &\quad +\sum_{0 < t' < 1} \underbrace{
            \kl
            {\q{\ttau{t'}}}
            {\pcond{\ttau{t'}} {\z{t}, \ttau{t}}}
        }_{\ltau} \\
        &\quad \underbrace{
            -\log \pcond
            {\z{0}, \ttau{0}}
            {\z{\dt}, \ttau{\dt}}
        }_{\loss_0}
    \Biggr].
\end{align}
Note that $\loss_1$ is a constant and can be ignored, and $\loss_0$ also becomes negligible when $\dt \to 0$. According to prior works~\citep{ho2020denoising, li2022diffusion}, the term $\lz$ can be simplified as
\begin{equation*}
    \lz = \lVert \zhat(\z{t}, \ttau{t}, t) - \z{0} \rVert^2.
\end{equation*}
We also add an anchor loss~\citep{gao-etal-2024-empowering} $\lossrm{anchor} = \mathbb{E}_q [-\log \p{y | \zhat(\z{t}, \ttau{t}, t)}]$ as a regularization term to avoid collapse of the embedding space.
Finally, the training objective of the proposed NeoDiff can be written as
\begin{equation*}
    \lours = \lz + \ltau + \lossrm{anchor}.
\end{equation*}

\newpage
\section{Experimental Settings}
\label{app:experimental_settings}
\subsection{Data Preprocessing}
We used byte pair encoding (BPE) \cite{sennrich-etal-2016-neural} for tokenization. Unlike previous work, we did not employ knowledge distillation \cite{kim2016sequence} for preprocessing to evaluate our model's performance under more challenging conditions.

\subsection{Model Configuration}
For our experiments, we set the maximum noise state $\smax$ to 100 and used the \emph{sqrt} schedule for training, optimized schedule for inference. To enhance model performance, we applied self-conditioning \cite{chen2022analog}. The transition schedule coefficient $k$ was set to 2, and the maximum truncation time $t_{\mathrm{max}}$ was set to 0.99. Following \citet{gao-etal-2024-empowering}, we also employed noise rescaling with a degradation score threshold $\dgsmax$ of 0.2. 

Regarding the model architecture, we adopted the configuration from \citet{gao-etal-2024-empowering} for the IWSLT14 De-En, WMT14 En-De, and WMT16 En-Ro datasets. For the QQP, Wiki-Auto, and QT datasets, we used the configuration from \citet{gong2022diffuseq} to enable a fair comparison with these models. Detailed settings are presented in Table~\ref{tab:settings}.

\subsection{Training and Generation}
We trained our models using NVIDIA RTX 3090 24G GPUs on Ubuntu 18.04 with FairSeq 0.12\cite{ott2019fairseq}(MIT-licensed). For the WMT14 En-De and WMT16 En-Ro datasets, training took nearly 4 days and 2 days, respectively, using 4 GPUs. For the IWSLT14 De-En dataset, training took approximately 1 day using a single GPU. The QQP, Wiki-Auto, and QT datasets each required around 8 hours of training on a single GPU. The training data splits are presented in Table~\ref{tab:dataset}.

During generation, we used 20 iteration steps ($K = 20$) without early stopping for the IWSLT14 De-En dataset. For the other datasets, we employed 10 iteration steps without early stopping, which is faster than the 20 steps ($k = 20$) used by \citet{gao-etal-2024-empowering} across all datasets. We utilized 2D parallel decoding and selected the best sentence using the minimum Bayes risk (MBR) \cite{kumar-byrne-2004-minimum} method based on the BLEU score. The reported results are averaged over 3 runs. The random seed is set to 7.

\begin{table*}[h]

\centering
\begin{tabular}{lccc}
\toprule
Models       & $K$    & Speed (sentences/second) & Memory Cost (MB) \\
\midrule
Transformer* & n    & 6.05                     & -                \\
CMLM*        & 10   & 11.80                    & -                \\
DiffuSeq*    & 2000 & 0.06                     & -                \\
SeqDiffuSeq* & 2000 & 0.05                     & -                \\
Difformer    & 20   & 6.49                     & 2034             \\
NeoDiff      & 20   & 5.12                     & 2080             \\
\bottomrule
\end{tabular}
\caption{Runtime Comparison on IWSLT14 De-En. *: Results from \citet{gao-etal-2024-empowering}. Others are reproduced. }
\label{tab:runtime_comparison}

\end{table*}

\subsection{Optimized Extrinsic Time Schedule}
\label{appendix:time_schedule}

We propose a systematic approach to optimize the extrinsic time schedule $\mathbf{S} = \{t_1, t_2, ..., t_K\}$, where $K$ denotes the number of diffusion steps and $t_i \in [0,1]$ with $t_1 < t_2 < ... < t_K$. While previous works~\citep{dhariwal2021diffusion,chen2023importancenoiseschedulingdiffusion} focus on optimizing noise schedules with fixed time steps, we directly optimize the time schedule through Bayesian optimization. Our method extends~\citet{li-etal-2024-shot-temporal}'s framework from discrete subset selection to continuous optimization over the complete schedule.

At its core, our approach is straightforward: we sample text using different time schedules on the validation set and select the schedule that achieves the highest BLEU score for inference. The optimization process (Algorithm~\ref{alg:temporal-pruning}) employs Gaussian Process-based Bayesian optimization with the GP-Hedge acquisition function~\citep{brochu2011portfolio}. Starting from a uniform time schedule, we iteratively propose candidate schedules using Limited-memory BFGS~\cite{Liu1989OnTL} and evaluate them using BLEU scores on the validation set. This approach enables precise calibration of the time schedule while maintaining the ordering constraint $t_1 < t_2 < ... < t_K$. Following~\citet{li-etal-2024-shot-temporal}, we limit optimization to 100 iterations, keeping the computational overhead negligible compared to model training time\cite{li-etal-2024-shot-temporal}. The resulting task-specific schedules demonstrate improved generation quality while maintaining computational efficiency.

\begin{algorithm*}[h]
\caption{Extrinsic Time Schedule Calibration via Bayesian Optimization}
\label{alg:temporal-pruning}
\begin{algorithmic}[1]
\REQUIRE Trained Diffusion Model $M$, \\
Initial Extrinsic Time Schedule $\mathbf{S}_{\mathrm{init}}=\{t_1,t_2,...,t_K\}$, where $t_i \in \mathbb{R}$ and $0 \leq t_1 < t_2 < ... < t_K \leq 1$, \\
Optimization iterations $n_{\mathrm{iter}}$, \\
Domain for elements in Extrinsic Time Schedule $\mathcal{D} \subset [0, 1]$, \\
Source Text $T_{\mathrm{src}}$, \\
Target Text $T_{\mathrm{tgt}}$
\ENSURE Optimized Extrinsic Time Schedule $\mathbf{S}_{\mathrm{opt}} = \{t'_1,t'_2,...,t'_K\}$, where $t'_i \in \mathbb{R}$ and $0 \leq t'_1 < t'_2 < ... < t'_K \leq 1$

\STATE Initialize $\mathbf{S}_{\mathrm{init}}=\{t_1,t_2,...,t_K\}$ such that $t_i$ are uniformly spaced in $[0, 1]$.
\STATE Perform a sampling on $T_{\mathrm{src}}$ using diffusion model $M$ and extrinsic time schedule $\mathbf{S}_{\mathrm{init}}$, yielding predicted text $T_{\mathrm{pred}}$.
\STATE Compute the BLEU score $\mathrm{BLEU}(T_{\mathrm{tgt}},T_{\mathrm{pred}})$ using $T_{\mathrm{tgt}}$ and $T_{\mathrm{pred}}$.
\STATE Initialize the observation set for Bayesian optimization: $O \gets \{ (\mathbf{S}_{\mathrm{init}}, \mathrm{BLEU}(T_{\mathrm{tgt}},T_{\mathrm{pred}})) \}$.

\FOR{$i = 1$ to $n_{\mathrm{iter}}$}
\STATE Update the Gaussian Process posterior given observations $O$.
\STATE Generate a candidate set $\mathcal{D}' = \{\mathbf{S}'_1, \mathbf{S}'_2, ..., \mathbf{S}'_N\}$, where each $\mathbf{S}'_j = \{t'_{j1}, t'_{j2}, ..., t'_{jK}\}$ represents a candidate extrinsic time schedule with $t'_{jk} \in \mathcal{D}$ and $0 \leq t'_{j1} < t'_{j2} < ... < t'_{jK} \leq 1$. The candidate set $\mathcal{D}'$ is generated by performing 20 iterations of Limited-memory BFGS~\cite{Liu1989OnTL} with 5 random initial points within $\mathcal{D}^K$.
\STATE Compute the acquisition function value $\alpha_{\text{GP-Hedge}}(\mathbf{S}'_j)$~\cite{brochu2011portfolio} for all $\mathbf{S}'_j \in \mathcal{D}'$.
\STATE Select the next observation point $\mathbf{S}_i = \arg\max_{\mathbf{S}'_j \in \mathcal{D}'} \alpha_{\text{GP-Hedge}}(\mathbf{S}'_j)$.
\STATE Perform a sampling on $T_{\mathrm{src}}$ using $M$ and $\mathbf{S}_i$, yielding predicted text $T'_{\mathrm{pred}}$.
\STATE Compute the BLEU score $\mathrm{BLEU}(T_{\mathrm{tgt}},T'_{\mathrm{pred}})$ using $T_{\mathrm{tgt}}$ and $T'_{\mathrm{pred}}$.
\STATE Update the observation set: $O \gets O \cup \{ (\mathbf{S}_i, \mathrm{BLEU}(T_{\mathrm{tgt}},T'_{\mathrm{pred}})) \}$.
\ENDFOR

\STATE $\mathbf{S}_{\mathrm{opt}} = \arg\max_{(\mathbf{S}, \mathrm{BLEU}) \in O} \mathrm{BLEU}$

\end{algorithmic}
\end{algorithm*}

\section{Additional Diversity Analysis on QQP dataset}
\label{app:additional_qqp_diversity_analysis}

In this section, we provide a detailed comparison of NeoDiff and Transformer on the QQP task, specifically focusing on multi-candidate generation and inter-sentence diversity. The results presented in Table~\ref{tab:appendix_qqp_diversity_comparison} complement the main paper's Table~\ref{tab:llm_evaluation_combined} by offering a deeper look into how diversity metrics evolve with an increasing number of generated samples ($b$).

\begin{table*}[t!] 
\centering

\small 
\begin{tabular}{@{}lcccccc@{}}
\toprule
\textbf{Model} & \textbf{b} & \textbf{Semantic Faithfulness} & \textbf{Fluency} & \textbf{Completeness} & \textbf{Phrasing Diversity} & \textbf{Inter-Sentence div-4} \\
\midrule
Transformer & 1  & 83.64          & 92.56          & 84.96          & \textbf{57.19} & \textbf{1.000} \\
Transformer & 5  & 83.70          & \textbf{94.73} & 86.05          & 54.52          & 0.686 \\
Transformer & 10 & 83.93          & 94.64          & 86.02          & 54.55          & 0.561 \\
\midrule
NeoDiff     & 1  & 84.24          & 88.95          & 87.83          & 39.18          & \textbf{1.000} \\
NeoDiff     & 5  & 85.63          & 90.69          & 88.39          & 41.62          & 0.684 \\
NeoDiff     & 10 & \textbf{87.42} & 91.87          & \textbf{88.79} & 45.83          & 0.631 \\
\bottomrule
\end{tabular}
\caption{Detailed comparison of NeoDiff and Transformer on the QQP task. Metrics include Semantic Faithfulness, Fluency, Completeness, Phrasing Diversity (single-sample), and Inter-Sentence Diversity (Inter-Sentence Div-4, multi-candidate).}
\label{tab:appendix_qqp_diversity_comparison} 
\end{table*}

\section{Efficiency Analysis}
\label{app:efficiency}

\subsection{Bayesian Optimization Overhead (WMT14 En-De):}
\label{app:bayesian}
\begin{itemize}
    \item Training: 505.88 RTX3090 GPU Hours
    \item Bayesian Optimization: 28.1 RTX3090 GPU Hours (approximately 6\% of training time)
\end{itemize}

\textbf{Note:} The cost of Bayesian optimization is directly proportional to the amount of data sampled in each iteration. While we used the entire WMT14 validation set, significantly reducing the sample size (e.g., to 20 samples) can drastically lower this overhead to less than 0.1 GPU Hours~\citep{li-etal-2024-shot-temporal}.

\subsection{Runtime Comparison (IWSLT14 De-En)}
\label{sec:runtime}

Table~\ref{tab:runtime_comparison} presents a runtime comparison of NeoDiff and several baselines on the IWSLT14 De-En dataset. We measured inference speed (sentences/second) and memory cost (MB).  NeoDiff demonstrates competitive inference speed, processing 5.12 sentences per second, which is comparable to Difformer's 6.49 sentences per second. While significantly faster than diffusion-based models like DiffuSeq and SeqDiffuSeq, NeoDiff's speed is lower than the highly optimized Transformer and CMLM models.  In terms of memory usage, NeoDiff's 2080 MB consumption is similar to Difformer's 2034 MB.

\section{Step-wise Generation Examples on IWSLT14 De-En for NeoDiff and Difformer}
\label{sec:appendix_translation_examples}

Table~\ref{tab:neodiff_vs_difformer_1} and \ref{tab:neodiff_vs_difformer_2} present a detailed comparison of the translation generation process on IWSLT14 De-En dataset between NeoDiff and Difformer(continuous diffusion model). After incorporating the three aforementioned components(Poisson process, time predictor and optimized schedule), NeoDiff demonstrates more accurate and faster convergence in translation on some sentences compared to Continuous Diffusion Model(Difformer), as illustrated by the step-by-step generation process. Specifically, NeoDiff avoids some of the common pitfalls of diffusion models, such as getting stuck in local optima or generating repetitive phrases.

\section{Fine-grained Controlled Generation through Token Manipulation}\label{appendix:token_examples}

We demonstrate NeoDiff's capability for token-level controlled generation while preserving semantic consistency across translations. Given a source sentence $\mathbf{x}_\text{src}$ and its latent representation $\mathbf{z}_0$, we replace a single token to obtain a modified source $\mathbf{x}'_\text{src}$. For translation, we initialize the process with $\mathbf{z}_0$ and set $\tau=1$ only for the modified token position, maximizing noise specifically at that location. This targeted noise application enables precise semantic modifications in the output translation $\mathbf{x}_\text{tgt}'$ while preserving the remaining content. As shown in Table~\ref{tab:neodiff_more_cases}, NeoDiff achieves localized modifications, whereas baseline methods like Difformer tend to alter substantial portions of the output sentence. This controlled generation capability stems from our fine-grained noise paradigm, enabling token-specific manipulation of the generation process.

\begin{table*}[t]
    \centering
    \small
    \begin{tabular}{l*{6}{C{4.5em}}}
        \toprule
        \textbf{Hyper-parameters} &
        \textbf{WMT14 En-De} &
        \textbf{WMT16 En-Ro} &
        \textbf{IWSLT14 De-En} &
        \textbf{QQP} &
        \textbf{Wiki-Auto} &
        \textbf{QT} \\
        \midrule
        
        \textbf{Architecture} \\
        
        $d_{\mathrm{model}}$ &
        $512$ & $512$ & $512$ &
        $768$ & $768$ & $768$ \\

        $d_{\mathrm{emb}}$ &
        $128$ & $128$ & $128$ &
        $128$ & $128$ & $128$ \\

        $d_{\mathrm{ffn}}$ &
        $2048$ & $2048$ & $1024$ &
        $3072$ & $3072$ & $3072$ \\

        Heads &
        $8$ & $8$ & $4$ &
        $12$ & $12$ & $12$ \\

        Encoder Layers &
        $6$ & $6$ & $6$ &
        $6$ & $6$ & $6$ \\

        Decoder Layers &
        $6$ & $6$ & $6$ &
        $6$ & $6$ & $6$ \\

        Time Predictor Layers &
        $3$ & $1$ & $1$ &
        $1$ & $1$ & $1$ \\

        Activation &
        ReLU & ReLU & ReLU &
        ReLU & ReLU & ReLU \\
        \midrule

        \textbf{Diffusion} \\

        Steps &
        $10$ & $10$ & $20$ &
        $10$ & $10$ & $10$ \\

        Training Schedule &
        \emph{sqrt} & \emph{sqrt} & \emph{sqrt} &
        \emph{sqrt} & \emph{sqrt} & \emph{sqrt} \\

        Inference Schedule &
        Optimized & Optimized & Optimized &
        Optimized & Optimized & Optimized \\
        $\dgsmax$ &
        $0.2$ & $0.2$ & $0.2$ &
        $0.2$ & $0.2$ & $0.2$ \\

        Self-Conditioning &
        \checkmark & \checkmark & \checkmark &
        \checkmark & \checkmark & \checkmark \\
        \midrule

        \textbf{Training} \\
        Steps &
        $600$K & $400$K & $300$K &
        $50$K & $100$K & $100$K \\

        Batch Size (Tokens) &
        $32$K & $24$K & $8$K &
        $8$K & $12$K & $16$K \\

        Optimizer &
        AdamW & AdamW & AdamW &
        AdamW & AdamW & AdamW \\

        Adam $\beta$ &
        $(0.9, 0.98)$ & $(0.9, 0.98)$ &
        $(0.9, 0.98)$ & $(0.9, 0.98)$ &
        $(0.9, 0.98)$ & $(0.9, 0.98)$ \\

        Weight Decay &
        $0.01$ & $0.01$ & $0.01$ &
        $0.01$ & $0.01$ & $0.01$ \\

        Learning Rate &
        $5 \times 10^{-4}$ & $5 \times 10^{-4}$ &
        $5 \times 10^{-4}$ &
        $5 \times 10^{-4}$ & $2.3 \times 10^{-4}$ &
        $2 \times 10^{-4}$ \\
        
        Warmup &
        $10$K & $10$K & $10$K &
        $10$K & $10$K & $10$K \\

        Clip Gradient &
        $1.0$ & $1.0$ & $1.0$ &
        $1.0$ & $1.0$ & $1.0$ \\

        Dropout &
        $0.1$ & $0.1$ & $0.3$ &
        $0.1$ & $0.1$ & $0.1$ \\

        Length Predict Factor &
        $0.1$ & $0.1$ & $0.1$ &
        $0.1$ & $0.1$ & $0.1$ \\

        Label Smoothing &
        $0.1$ & $0.1$ & $0.1$ &
        $0.1$ & $0.1$ & $0.1$ \\
        \midrule

        \textbf{Inference} \\
        Steps &
        $10$ & $10$ & $20$ &
        $10$ & $10$ & $10$ \\

        Bayesian Optimization Rounds &
        $100$ & $100$ & $100$ &
        $100$ & $100$ & $100$ \\
        
        \bottomrule
    \end{tabular}
    \caption{The model architectures and hyper-parameters used in our experiments.}
    \label{tab:settings}
\end{table*}

\begin{figure*}[t]
\begin{mdframed}[backgroundcolor=gray!10]
\begin{verbatim}
Evaluate this translation from {src_lang} to {tgt_lang} (0-100 score):  
[Source] {source}  
[Reference] {reference}  
[Translation] {translation}  
Score these aspects STRICTLY IN THIS ORDER:
1. **Accuracy**: Faithfulness to source meaning  
2. **Fluency**: Naturalness in target language  
3. **Completeness**: Information retention  
4. **Creativity**: Handling of ambiguous or open-ended source content  

Return ONLY 4 numbers separated by commas, NO text. 
\end{verbatim}
\end{mdframed}

\vspace{1em}

\begin{mdframed}[backgroundcolor=gray!10]
\begin{verbatim}
Evaluate this paraphrase generation (0-100 score):  
[Original] {source}  
[Reference] {reference}  
[Paraphrase] {paraphrase}  
Score these aspects STRICTLY IN THIS ORDER:
1. **Semantic Faithfulness**: Meaning preservation from original
2. **Fluency**: Naturalness in language
3. **Completeness**: Retention of all information
4. **Phrasing Diversity**: Variation in wording/structure while preserving meaning

Return ONLY 4 numbers separated by commas, NO text. 
\end{verbatim}
\end{mdframed}

\caption{Prompt templates used for LLM-based evaluation. Top: Translation evaluation prompt. Bottom: Paraphrase evaluation prompt.}
\label{fig:prompts}
\end{figure*}

\begin{table*}[htb]
    \small
    \centering
    
    \begin{tabular}{ll}
        \toprule
        
    \texttt{\textbf{<src>}} &
        \texttt{und die welt in der wir jetzt leben sieht \colorbox{ocean}{so} aus .} \\

        \texttt{\textbf{<tgt>}} &
        \texttt{and the world we now live in looks \colorbox{ocean}{like this} .} \\
        \texttt{\textbf{<src'>}}  &
        \texttt{und die welt in der wir jetzt leben sieht \colorbox{change}{anders} aus .} \\

        \texttt{\textbf{<tgt'>}}  &
        \texttt{and the world we now live in looks \colorbox{change}{different} .} \\
        \midrule
        \textbf{Model} & \multicolumn{1}{c}{\textbf{Generated Content}} \\
        \midrule
        NeoDiff \texttt{\textbf{<tgt\_pred>}} &
        \texttt{and the world we live in now , looks \colorbox{ocean}{like this} .} \\
        NeoDiff \texttt{\textbf{<tgt'\_pred>}} &
        \texttt{and the world we live in now , looks \colorbox{change}{different} .} \\    
        \midrule

        Difformer \texttt{\textbf{<tgt\_pred>}}&
        \texttt{and the world we're living in now looks \colorbox{ocean}{like this} .} \\   
        Difformer \texttt{\textbf{<tgt'\_pred>}}&
        \texttt{and the world that we're living in \colorbox{change}{right now} , \colorbox{change}{it looks different}} . \\   

\toprule
    \texttt{\textbf{<src>}} &
        \texttt{sein ganzer arbeitsprozess hat sich  \colorbox{ocean}{danach} geändert .} \\

        \texttt{\textbf{<tgt>}} &
        \texttt{and his whole work process changed \colorbox{ocean}{after} that .} \\
        \texttt{\textbf{<src'>}}  &
        \texttt{sein ganzer arbeitsprozess hat sich \colorbox{change}{davon} geändert.} \\

        \texttt{\textbf{<tgt'>}}  &
        \texttt{His whole work process changed \colorbox{change}{because of} that.} \\
        \midrule
        \textbf{Model} & \multicolumn{1}{c}{\textbf{Generated Content}} \\
        \midrule
        NeoDiff \texttt{\textbf{<tgt\_pred>}} &
        \texttt{his whole work process has changed \colorbox{ocean}{after} that .} \\
        NeoDiff \texttt{\textbf{<tgt'\_pred>}} &
        \texttt{his whole work process has changed \colorbox{change}{from} that .} \\    
        \midrule

        Difformer \texttt{\textbf{<tgt\_pred>}}&
        \texttt{and his whole work process has changed \colorbox{ocean}{after} that .} \\   
        Difformer \texttt{\textbf{<tgt'\_pred>}}&
 
        \texttt{and his whole work process has changed \colorbox{change}{from this .}} \\   
\toprule
    \texttt{\textbf{<src>}} &
        \texttt{der zweite faktor sind die dienste , die wir \colorbox{ocean}{nutzen} .} \\

        \texttt{\textbf{<tgt>}} &
        \texttt{the second factor is the services we \colorbox{ocean}{use} .} \\
        \texttt{\textbf{<src'>}}  &
        \texttt{der zweite faktor sind die dienste , die wir \colorbox{change}{kennen}.} \\

        \texttt{\textbf{<tgt'>}}  &
        \texttt{The second factor is the services we \colorbox{change}{know}.} \\
        \midrule
        \textbf{Model} & \multicolumn{1}{c}{\textbf{Generated Content}} \\
        \midrule
        NeoDiff \texttt{\textbf{<tgt\_pred>}} &
        \texttt{the second factor is the services that we \colorbox{ocean}{use} .} \\
        NeoDiff \texttt{\textbf{<tgt'\_pred>}} &
        \texttt{the second factor is the services we \colorbox{ocean}{know} .} \\    
        \midrule

        Difformer \texttt{\textbf{<tgt\_pred>}}&
        \texttt{the second factor is the services that we \colorbox{ocean}{use} .} \\   
        Difformer \texttt{\textbf{<tgt'\_pred>}}&

        \texttt{the second factor is \colorbox{change}{really} the services that we \colorbox{change}{meet} .} \\   
        \bottomrule
    \end{tabular}
    \caption{Token manipulation example. }
    \label{tab:neodiff_more_cases}
    
\end{table*}

\begin{table*}[t]
\small
\centering
\begin{tabular}{p{3.5cm}p{3.5cm}p{3.5cm}p{3.5cm}}
\toprule
\textbf{Source} & \textbf{Reference} & \textbf{Base Translation} & \textbf{+P Translation} \\
\midrule
nein war nie eine möglichkeit gewesen . & no had never been an option . & no one has never been a possibility. & no had never been an opportunity. \\
\midrule
weiß jemand , was drei sekunden sind ? & does anyone know what three seconds are ? & does anybody know what three seconds? & does anyone know what three seconds are? \\
\midrule
sie hatten ein konzept von blauem blut . & they had a concept of blue blood . & they had a idea of blue blood. & they had a concept of blue blood. \\
\midrule
und raten sie was wir in dem angriffscode gefunden haben ? & and guess what we found in the attack code ? & and do you guess what we've found in the code of attack? & and guess what we found in the code of attack? \\
\midrule
jetzt sehen sie den dalmatiner . & now you see the dalmatian . & now this is the dalmatinan. & now you see the dalmatiner. \\
\midrule
denn die kategorien sagen mir , wie ich sie auseinander halten kann . & because the categories tell me how to tell them apart . & because the categories are telling me how i can keep it apart. & because the categories tell me how to keep them apart. \\
\midrule
wie konnte es möglich sein , dass wir dies tun ? & how could it be possible that we would do this ? & so how could it possible for us to do this? & how could it be possible that we could do this? \\
\midrule
aber es gab immer einen lebenszyklus in ihren präsentationen . & but there was always a life cycle to their presentations . & but there has always been a life cycle in your presentations. & but there was always a life cycle in their presentations. \\
\midrule
wir reden zwiespältig davon . & we talk about it ambivalently . & we're talking about it in elessly. & we talk about it continally. \\
\midrule
was geschah also jahre danach ? & so what happened years afterward ? & so for years after that, what happened? & so what happened years after that? \\
\bottomrule
\end{tabular}
\caption{Additional examples showing the improvements from introducing the Poisson diffusion process on IWSLT14 De-En dataset. The Base model often produces unnatural word ordering and incorrect lexical choices, while +P shows better handling of complex phrases and more natural English constructions.}
\label{tab:appendix_poisson}
\end{table*}

\begin{table*}[t]
\small
\centering
\begin{tabular}{p{3.5cm}p{3.5cm}p{3.5cm}p{3.5cm}}
\toprule
\textbf{Source} & \textbf{Reference} & \textbf{+P Translation} & \textbf{+PT Translation} \\
\midrule
sie haben ihr telefon gemietet . sie haben es nicht gekauft . & you rented your phone . you didn't buy it . & they've rtended your phone. they didn't buy it. & you rented your phone. you didn't buy it. \\
\midrule
ihre familie versammelte sich . & and the family gathered . & her family. & her family gathered. \\
\midrule
dunkler urin . dunkel . & dark urine . dark . & dark up. dark. & dark urine. dark. \\
\midrule
diese leute verdienen geld . & these guys make money . & these people are earking money. & these people make money. \\
\midrule
er ist ganz glücklich darüber , weil er sie getäuscht hat . & he'll be very happy because he's deceived you . & he's very happy about it because he decaked her. & he's very happy about it because he deceied her. \\
\midrule
er hatte 20 minuten herrlicher musik gehabt . & he had had 20 minutes of glorious music . & he'd had 20 minutes of god. & he'd had 20 minutes of glorious music. \\
\midrule
... es dem crowdsourcing beachtung schenkt . & ... paying attention to crowd-sourcing . & ... it's adghting to the crowdsourcing. & ... it gives attention to the crowdsourcing. \\
\midrule
er zeigte immer hier hin . & he kept pointing here . & he always showed here. & he always pointed over here. \\
\midrule
wenn man es verallgemeinert , passiert folgendes . & if you generalize this , something like this happens . & when you generate it, this is what happens. & when you generalize it, this is what happens. \\
\midrule
man konnte manhattan sehen . & you could see manhattan . & see manhattan. & you could see manhattan. \\
\bottomrule
\end{tabular}
\caption{Additional examples demonstrating the impact of the time predictor module on IWSLT14 De-En dataset. The examples show how the time predictor enables finer-grained control primarily through word substitutions and better token-level refinements by leveraging information from less-noised tokens to guide the denoising process.}
\label{tab:appendix_timepredictor}
\end{table*}

\begin{table*}[t]
\small
\centering
\begin{tabular}{p{3.5cm}p{3.5cm}p{3.5cm}p{3.5cm}}
\toprule
\textbf{Source} & \textbf{Reference} & \textbf{+PT Translation} & \textbf{Full Translation} \\
\midrule
so wie es früher eben entsprechend auf dem dorf passierte . & just like it used to happen in the village . & in the same way that happened in the village, it just happened. & just as it used to happen in the village. \\
\midrule
darum helfen sie da mit , fragen sie bei den leuten mal nach . & so you're helping out there , just ask the people . & so you can help with there, ask about people. & that's why they help there with, ask people to ask. \\
\midrule
noch immer sind wir dem storytelling als informationsvermittlung sehr , sehr stark verhaftet . & what's left is storytelling . & but we're still very sted to storytelling as an information reation, very vivily arrested. & we're still very to storytelling as an information mediation, very, very arrested. \\
\midrule
wir wählen jedes jahr einige fellows aus und wir lassen sie mit stadtverwaltungen arbeiten . & we select a few fellows every year and we have them work with city governments . & we choose some fellows every year, and we have them work with city adminicies. & we choose some fellows every year, and we let them work with urban management. \\
\midrule
also bot ich einen 10000 \$ preis an software für die gewinner . & so i offered a 10,000 dollar prize of software to the winning team . & so i offered a \$10,000 price for the winner software. & so i offered a 100,000 price of software for the winners. \\
\midrule
dies ist in unserem ganzen land der zweitgrösste abfallfluss amerikas . & this , all over the country , is the second largest waste stream in america . & this is in our entire country, the two-largest waste river of america. & this is the second est waste flow in america's land in our entire country. \\
\midrule
wir haben eine art gleichgewicht erreicht . & we have reached a kind of equipoise . & we've reachved some kind of equilibrium. & we've reached some kind of balance. \\
\midrule
und das ist aber ganz im anfang . & and that's just the beginning . & and that's just at the very beginning. & and that's at the very beginning. \\
\midrule
ich bin überzeugt , dass man irgendwie zur nostalgie , zu wunschdenken hingezogen ist . & i'm convinced that there's some sort of pull to nostalgia , to wishful thinking . & i'm convinced you've been drawn to nostalgia, sort of wokkthinking. & i'm believe that there's kind of moved to nostalgia, you're moved to thinking. \\
\midrule
wir haben uns daran gewöhnt , dass dinge linear passieren . & we no longer imagine the thing in images things in images , but codify them through language . & we were used to make things happen to linear. & so we've been used to linear that things happen. \\
\bottomrule
\end{tabular}
\caption{Additional examples showing the impact of the optimized schedule on IWSLT14 De-En dataset. These examples demonstrate how the schedule primarily influences the overall sampling trajectory at the sentence level, leading to more natural sentence constructions and better semantic coherence.}
\label{tab:appendix_schedule}
\end{table*}

\begin{table*}[t]
\small
\centering
\begin{tabular}{p{1.5cm}|p{5.5cm}|p{5.5cm}}
\toprule
\textbf{Time Step} & \textbf{Difformer Translation} & \textbf{NeoDiff Translation (Ours)} \\ \midrule
\multicolumn{3}{l}{\textbf{Source:} ihr problem ist , dass sie zu wenig haben .} \\
\multicolumn{3}{l}{\textbf{Reference:} their problem is that they have too little .} \\ \midrule
0           & dete@@ social tious falsche foot ere security madeupword0000 sorry fold says write chri@@ & 28 lar@@ electricity terms surface ting madeupword0001 madeupword0000 ® gen ﻿         \\
1           & your problem is that is that they have too little .                                       & their problem is they they have too little .                                           \\
2           & the problem of that is that they have too little .                                      & their problem is that they have too little .                                           \\
3           & the problem of that is that they have too little .                                      & their problem is that they have too little .                                           \\
4           & the problem your you is that they have too little .                                     & their problem is that they have too little .                                           \\
5           & the problem your problem is that they have too little .                                  & their problem is that they have too little .                                           \\
6           & and , your problem is that they have too little .                                       & their problem is that they have too little .                                           \\
7           & now , your problem is that they have too little .                                      & their problem is that they have too little .                                           \\
8           & now , your problem is , they have too little .                                        & their problem is that they have too little .                                           \\
9           & now , your problem is , you have too little .                                         & their problem is that they have too little .                                           \\
...          & ... & ... \\
20          & now , your problem is , you have too little .                                      & their problem is that they have too little .                                           \\ \midrule
\textbf{Final} & now , your problem is , you have too little .                         & their problem is that they have too little .                                           \\ \midrule
\multicolumn{3}{l}{} \\
\multicolumn{3}{l}{\textbf{Source:} denn die kategorien sagen mir , wie ich sie auseinander halten kann .} \\
\multicolumn{3}{l}{\textbf{Reference:} because the categories tell me how to tell them apart .} \\ \midrule
0           & es clay madeupword0002 ahead jobs involved line madeupword0001 fold <pad> <unk> giving bu@@ <unk> ers sa@@ & market@@ madeupword0003 madeupword0001 van mas price gun ba madeupword0000 <unk> 3 ator anima@@ once         \\
1           & because the categ@@ ories tell tell me how can can hold them apart . & because the categ@@ ories tell me how to keep them apart . \\
2           & because the categ@@ ories tell tell me how can can hold them apart . & because the categ@@ ories tell me how to keep them apart . \\
3           & because the categ@@ ories are tell me how i can hold it apart . & because the categ@@ ories tell me how to keep them apart . \\
4           & because the categ@@ ories are telling me how i can hold it apart . & because the categ@@ ories tell me how to keep them apart . \\
5           & because the categ@@ ories are telling me how i can hold it apart . & because the categ@@ ories tell me how to keep them apart . \\
6           & because the categ@@ ories are telling me how i can hold it apart . & because the categ@@ ories tell me how to keep them apart . \\
7           & because the categ@@ ories are telling me how i can hold it apart . & because the categ@@ ories tell me how to keep them apart . \\
8           & because the categ@@ ories are telling me how i can hold it apart . & because the categ@@ ories tell me how to keep them apart . \\
9           & because the categ@@ ories are telling me how i can hold it apart . & because the categ@@ ories tell me how to keep them apart . \\
...          & ... & ... \\
20          & because the categ@@ ories are telling me how i can keep it apart . & because the categ@@ ories tell me how to keep them apart .\\ \midrule
\textbf{Final} & because the categories are telling me how i can keep it apart . & because the categories tell me how to keep them apart . \\ \bottomrule
\end{tabular}
\caption{Step-by-step generation process of Difformer(continuous diffusion model) and NeoDiff on IWSLT14 De-En dataset(Part 1/2). NeoDiff converges to the correct translation more quickly and accurately.}
\label{tab:neodiff_vs_difformer_1}
\end{table*}

\begin{table*}[t]
\small
\centering
\begin{tabular}{p{1.5cm}|p{5.5cm}|p{5.5cm}}
\toprule
\textbf{Time Step} & \textbf{Difformer Translation} & \textbf{NeoDiff Translation (Ours)} \\ \midrule
\multicolumn{3}{l}{\textbf{Source:} ich sprach also einige monate später bei einer konferenz .} \\
\multicolumn{3}{l}{\textbf{Reference:} so i spoke at a conference a couple months after that .} \\ \midrule
0           & complex positive which affect o went attac@@ care@@ <pad> gers <pad> david fri@@ level & madeupword0001 le@@ leaders news madeupword0003 sta@@ cannot än@@ madeupword0001 spin@@ published mes@@ exhi@@        \\
1           & so i i to a few months later at a conference . & so i spoke at at a a few months later . \\
2           & so i i to a few months later at a conference . & so i spoke at a conference a few months later . \\
3           & so i i about a few months later at a conference . & so i spoke at a conference a few months later . \\
4           & so i i about a few months later at a conference . & so i spoke at a conference a few months later . \\
5           & so i i talking a few months later at a conference . & so i spoke at a conference a few months later . \\
6           & so i i talking a few months later at a conference . & so i spoke at a conference a few months later . \\
7           & so i i talking a few months later at a conference . & so i spoke at a conference a few months later . \\
8           & so i was talking a few months later at a conference . & so i spoke at a conference a few months later . \\
9           & so i was talking a few months later at a conference . & so i spoke at a conference a few months later . \\
...          & ... & ... \\
20          & so i was talking a few months later at a conference . & so i spoke at a conference a few months later . \\ \midrule
\textbf{Final} & so i was talking a few months later at a conference . & so i spoke at a conference a few months later . \\ \bottomrule
\end{tabular}
\caption{Step-by-step generation process of Difformer(continuous diffusion model) and NeoDiff on IWSLT14 De-En dataset(Part 2/2). NeoDiff converges to the correct translation more quickly and accurately.}
\label{tab:neodiff_vs_difformer_2}
\end{table*}

\begin{table*}[t]
    \small
    \centering

    \begin{tabular}{l*{6}{C{4em}}}
        \toprule
        \textbf{Splits} &
        \textbf{WMT14 En-De} &
        \textbf{WMT16 En-Ro} &
        \textbf{IWSLT14 De-En} &
        \textbf{QQP} &
        \textbf{Wiki-Auto} &
        \textbf{QT} \\
        \midrule
        
        Training &
        $4,500,966$ & $608,319$ & $160,215$ &
        $144,715$ & $677,751$ & $116,953$ \\

        Validation &
        $3,000$ & $1,999$ & $7,282$ &
        $2,048$ & $2,048$ & $2,048$ \\

        Test &
        $3,003$ & $1,999$ & $6,750$ &
        $2,500$ & $5,000$ & $10,000$ \\
        \bottomrule
    \end{tabular}
    \caption{The dataset splits used in our experiments.}
    \label{tab:dataset}
\end{table*}

\end{document}